%% file: 00_main.tex
\documentclass[runningheads]{llncs}

\usepackage{eccv}

\usepackage{eccvabbrv}

\usepackage{graphicx}
\usepackage{booktabs}

\usepackage[accsupp]{axessibility}  

\usepackage{hyperref}

\usepackage{orcidlink}
\usepackage{caption}
\newsavebox{\ushapetab}
\usepackage{booktabs}
\usepackage{multirow}
\usepackage{makecell}
\usepackage{adjustbox}
\usepackage{verbatim}
\usepackage{placeins}
\usepackage{tabularx}
\usepackage{pifont}
\newcommand{\cmark}{\ding{51}}
\newcommand{\xmark}{\ding{55}}
\usepackage{wrapfig}
\usepackage{capt-of}  

\begin{document}

\title{DRIFT: Difficulty-aware Rectified Flows for Through-plane MRI Super-Resolution}

\titlerunning{DRIFT for Through-plane MRI SR}

\author{Yoonseok Choi\inst{1}\orcidlink{0009-0004-4807-1884} \and
Eun-Gyu Ha\inst{1}\orcidlink{0000-0002-4436-1751} \and
Daniel Kim\inst{1}\orcidlink{0009-0008-0615-5093} \and
Mohammed A. Al-masni\inst{2}\orcidlink{0000-0002-1548-965X} \and
Ming-Hsuan Yang\inst{3}\orcidlink{0000-0003-4848-2304} \and
Dong-Hyun Kim\inst{1}\orcidlink{0000-0002-6717-7770}\thanks{Corresponding author.}}

\authorrunning{Y.~Choi et al.}

\institute{Department of Electrical and Electronic Engineering, Yonsei University, Seoul, Republic of Korea \and
Department of Bioengineering, King Fahd University of Petroleum \& Minerals~(KFUPM), Dhahran, Saudi Arabia \and
Electrical Engineering and Computer Science, University of California at Merced, Merced, USA \\
\email{\{yoonseokchoi, eungyuha9722, danny4159, donghyunkim\}@yonsei.ac.kr}\\
\email{m.almasani@sejong.ac.kr}\\
\email{mhyang@ucmerced.edu}}

\maketitle
\input{section/00_abstract}
\input{section/01_intro}

\input{section/02_related_works}
\input{section/03_method}
\input{section/04_experiments}
\input{section/05_conclusion}

\section*{Acknowledgements}
This work was supported by the National Research Foundation of Korea (NRF) grant funded by the Korea government (MSIT) (RS-2025-00561616).

%
%
\clearpage
\bibliographystyle{splncs04}
\bibliography{main}

\end{document}


\title{Supplementary Material for DRIFT: Difficulty-aware Rectified Flows for Through-plane MRI Super-Resolution}
\titlerunning{Supplementary Material for DRIFT}

\author{Yoonseok Choi et al.}
\authorrunning{Y.~Choi et al.}
\institute{}

\maketitle

\input{supple}

\bibliographystyle{splncs04}
\bibliography{main}

%% file: section/00_abstract.tex
\begin{abstract}
Magnetic Resonance Imaging~(MRI) is often acquired with anisotropic resolution to reduce scan time, producing stair-step artifacts along the through-plane direction. 
In through-plane MRI super-resolution, an efficiency–fidelity trade-off arises: feed-forward regressors are fast but oversmooth at large slice-thicknesses, while sampling-based methods improve fidelity at high inference cost. 
We propose DRIFT, a two-stage thickness-conditioned rectified flow framework for through-plane MRI super-resolution with continuous input slice-thickness. 
Stage 1 employs an Anatomical Projection Network~(APN) to map low-resolution patches to a coarse high-resolution manifold, providing a deterministic anatomical initialization that shortens the residual transport of Stage 2 and stabilizes slice-wise refinement. 
Stage 2 refines details via rectified flow and introduces a Physics-Aware Difficulty~(PAD) metric derived from slice-thickness induced through-plane bandwidth deficit to guide an Adaptive Integration Scheduler~(AIS), allocating ODE steps by thickness. 
A Consistent Endpoint Trajectory Alignment~(CETA) loss enforces thickness-consistent reconstructions. 
Experiments show that DRIFT outperforms super-resolution baselines while reducing inference cost. 
Code, models, and interactive demos are available at \url{https://yoonseokchoi-ai.github.io/drift-eccv2026/}.
\keywords{Super-Resolution \and Rectified Flow \and MRI}
\end{abstract}

%% file: section/01_intro.tex
\section{Introduction}
\label{sec:intro}

High-resolution~(HR) isotropic Magnetic Resonance Imaging~(MRI) provides crucial anatomical details for precise clinical diagnosis and quantitative neuroanatomical analysis. However, fundamental physical constraints such as the signal-to-noise ratio~(SNR) and acquisition time often necessitate anisotropic acquisition, resulting in thick-slice volumes with low through-plane resolution. To bridge this gap, through-plane super-resolution~(SR) approaches have emerged as a vital task to reconstruct isotropic volumes from anisotropic inputs.

Various approaches address through-plane SR under anisotropic acquisition, including self-supervision from in-plane and through-plane asymmetry~\cite{zhao2020smore}, alias-aware regression for thick-slice artifacts~\cite{song2023afcm}, and synthesis-driven training for heterogeneous clinical protocols~\cite{iglesias2023synthsr}. While these pipelines mitigate stair-step artifacts and improve reconstruction quality, regression-style estimators often favor conservative predictions and tend to be optimized for discrete target scales or protocol settings, which can limit generalization under the continuous slice-thickness variation encountered in practice. More broadly, continuously varying slice-thickness motivates SR formulations that represent images as continuous functions rather than relying on discrete grid-to-grid regression.

Implicit Neural Representations~(INR) methods~\cite{chen2021liif,lee2022lte} represent images as continuous functions of spatial coordinates, with medical adaptations such as ArSSR~\cite{wu2022arssr} and SA-INR~\cite{wang2024sainr} extending this paradigm to 3D brain MRI. Despite their flexibility, these models face two challenges that matter for through-plane SR in MRI. First, they suffer from spectral bias~\cite{rahaman2019spectral}, where Multi-Layer Perceptron~(MLP)-based continuous functions prioritize low-frequency structural components, which tends to under-recover high-frequency textures essential for clinical fidelity. Second, most existing arbitrary-scale models are trained using isotropic downsampling, which may not reflect the unique physics of MRI slice acquisition, where degradation is dominated by slice-profile governed through-plane integration~\cite{peng2020saint,pauly1991slr}.

Generative approaches, particularly diffusion-based frameworks, have recently demonstrated superior performance in capturing complex textures. For instance, TPDM~\cite{lee2023tpdm} utilizes pre-trained perpendicular 2D diffusion models to improve 3D imaging quality. However, the prohibitive inference cost remains a significant bottleneck for real-time clinical use. While recent advancements such as ResShift~\cite{yue2023resshift} have reduced sampling steps by initiating the sampling process from the low-resolution~(LR) image instead of noise, they utilize a fixed-step scheduler regardless of the degradation severity of the input. In contrast, AdaDiffSR~\cite{fan2024adadiffsr} introduces region-aware dynamic acceleration, but its image-dependent perception modules introduce non-negligible computational overhead and remain fundamentally detached from the underlying physical degradation~(\ie, slice-thickness) inherent in MRI acquisition.

\begin{figure}[t]
\centerline{\includegraphics[width=\textwidth]{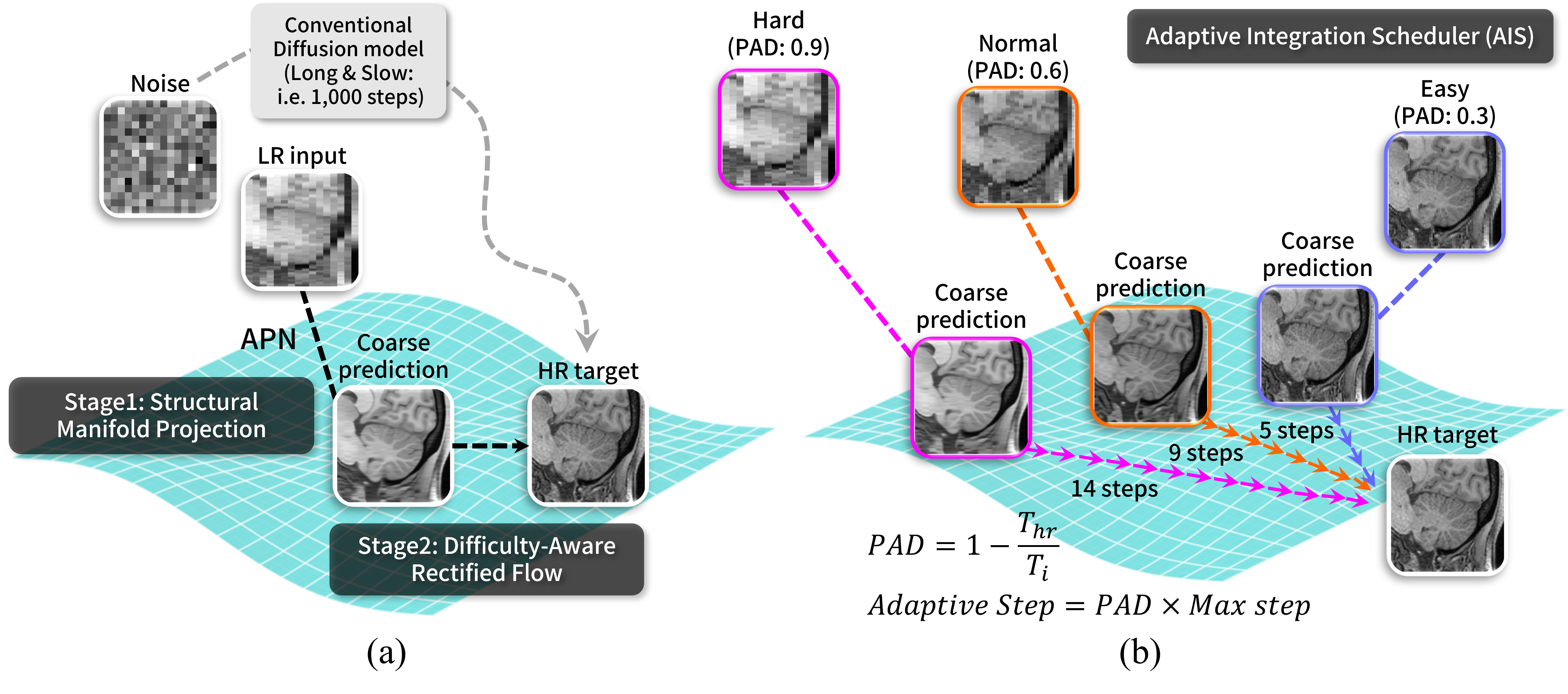}}
\caption{\textnormal{Conceptual overview of DRIFT inference.
(a) DRIFT initializes rectified flow from a coarse manifold estimate $z$ rather than noise, shortening the residual transport trajectory.
(b) AIS allocates ODE steps using the PAD metric, assigning more steps to thick-slice hard cases and fewer steps to thin-slice easy cases.}}
\label{fig:drift_concept}
\vspace{-5mm}
\end{figure}

\begin{figure}[t]
\centerline{\includegraphics[width=\textwidth]{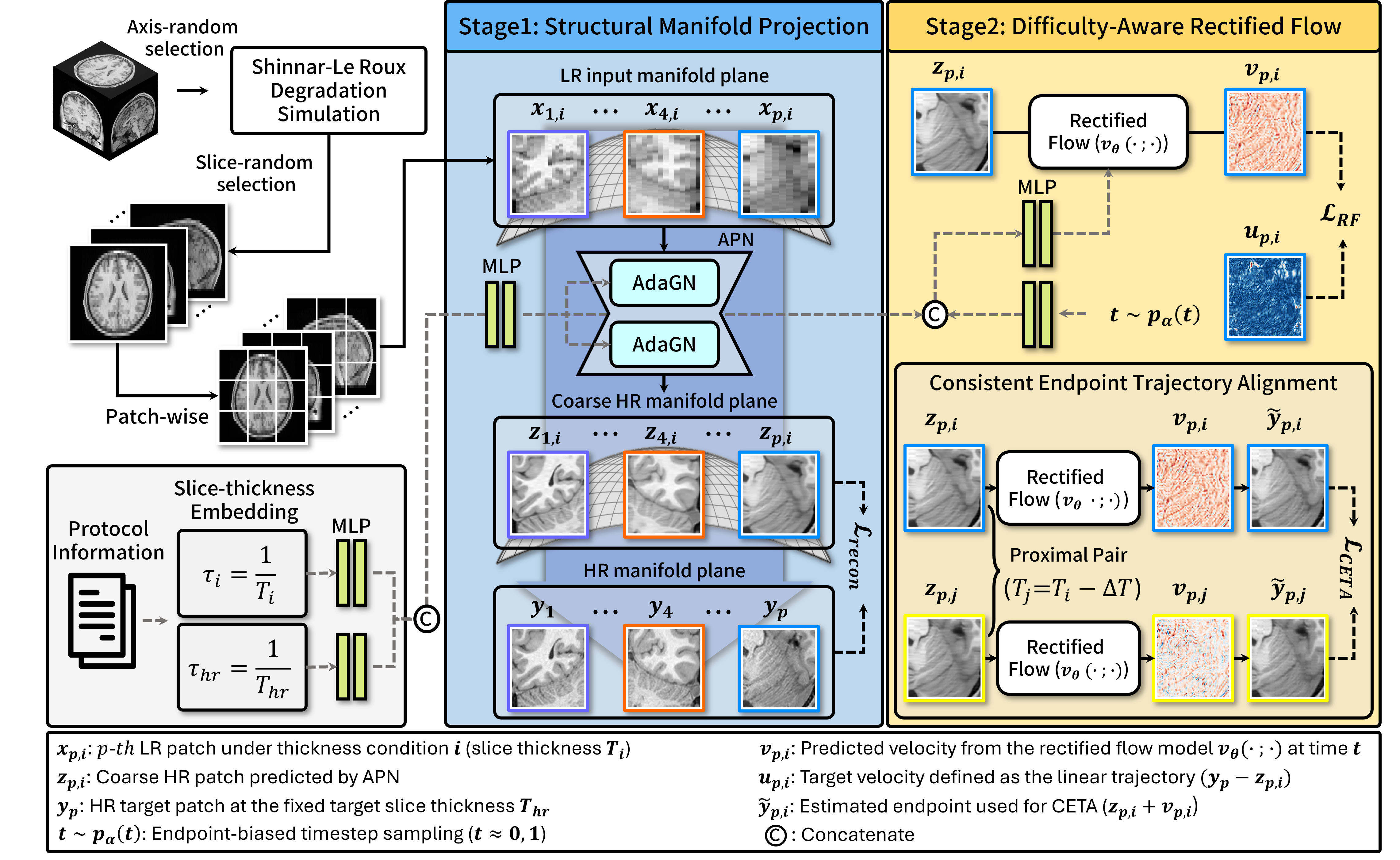}}
\caption{\textnormal{Training pipeline of DRIFT. Stage~1 uses an Anatomical Projection Network~(APN) to project LR patches to a coarse HR manifold conditioned on inverse thickness $\tau$. Stage~2 refines textures via rectified flow by predicting a straight-line velocity field. We enforce Consistent Endpoint Trajectory Alignment~(CETA) by regularizing trajectories between proximal thickness pairs~($T_j=T_i-\Delta T$, $\Delta T=1$\,mm). Notation is summarized in the figure legend.}} \label{fig:drift_training}
\vspace{-3mm}
\end{figure}

We propose \textbf{D}ifficulty-aware \textbf{R}ectif\textbf{I}ed \textbf{F}low for \textbf{T}hrough-plane SR~(DRIFT), a two-stage framework for reconstructing a specified target resolution from inputs with continuous slice-thicknesses~(Fig.~\ref{fig:drift_concept}). DRIFT conditions both stages on the input and target slice thicknesses, allowing a single model to adapt its reconstruction process across different through-plane degradation levels. In Stage~1, DRIFT uses an Anatomical Projection Network~(APN) to map LR patches to a coarse HR manifold, providing a structured initialization that shortens the subsequent refinement trajectory. In Stage~2, DRIFT refines high-frequency details using rectified flow, enabling deterministic inference with a reduced number of function evaluations~(NFEs)~\cite{liu2023flow,lipman2023flow}.
During training, DRIFT simulates thickness-diverse anisotropic inputs using a slice-profile model grounded in the Shinnar-Le Roux~(SLR) algorithm~\cite{pauly1991slr} which mirrors the slice-selection process in MRI acquisition. We sample slice-thicknesses from common clinical ranges~\cite{iglesias2021joint,iglesias2023synthsr} and degradation axis, and apply an SLR-derived slice profile along the chosen axis to model through-plane signal integration. We then resample to the thick-slice grid and map the result back to the HR grid with nearest-neighbor interpolation to reproduce stair-step artifact morphology. To encourage thickness-consistent outputs across conditions, we introduce the Consistent Endpoint Trajectory Alignment~(CETA) loss using proximal thickness pairs~(Fig.~\ref{fig:drift_training}).

At test time, DRIFT uses a Physics-Aware Difficulty~(PAD) metric computed from slice-thickness metadata to drive an Adaptive Integration Scheduler~(AIS), which allocates ordinary differential equation~(ODE) steps without image-dependent decision modules~\cite{fan2024adadiffsr}~(Fig.~\ref{fig:drift_concept}). Our contributions are summarized as follows:
\begin{itemize}
    \item We formulate input slice-thickness-continuous through-plane MRI SR as a two-stage anatomical projection-to-transport problem, where a thickness-conditioned APN provides deterministic and spatially correlated slice-wise initialization that shortens the residual rectified-flow trajectory.
    \item We introduce an MRI protocol-aware conditioning and inference mechanism: both stages are conditioned on input and target inverse thickness, while PAD/AIS maps slice-thickness-induced bandwidth deficit to adaptive ODE step budgets without auxiliary image-difficulty networks.
    \item We propose CETA, a proximal-thickness endpoint alignment loss that regularizes rectified-flow trajectories across neighboring slice-thickness conditions and improves over naive random-pair consistency.
    \item We validate DRIFT on public and real thick-slice MRI datasets with comprehensive reconstruction, ablation, standardized efficiency, zero-shot clinical transfer, and through-plane continuity analyses.
\end{itemize}
\vspace{-5mm}

%% file: section/02_related_works.tex
\section{Related Works}
\label{sec:related}

\textbf{Through-plane MRI SR.} 
Through-plane MRI SR focuses on reconstructing isotropic volumes from anisotropic thick-slice scans, where the resolution is significantly lower along the slice-selection direction. Self-supervised frameworks such as SMORE~\cite{zhao2020smore} and SIMPLE~\cite{benisty2025simple} exploit the resolution asymmetry between in-plane and through-plane directions to learn reconstruction from the high-resolution in-plane data itself. To address aliasing artifacts inherent in thick-slice acquisition, AFCM~\cite{song2023afcm} introduces an alias-free formulation using co-modulated generators. For clinically heterogeneous data, SynthSR~\cite{iglesias2023synthsr} and the joint optimization framework by Iglesias~\etal~\cite{iglesias2021joint} utilize synthetic data from segmentation labels to ensure robustness. More recent methods explore generative priors. TPDM~\cite{lee2023tpdm} leverages perpendicular 2D diffusion models for 3D reconstruction. Despite these advances, most methods assume discrete scale degradation models, failing to generalize across the continuous slice-thickness manifold encountered in clinical practice.

\noindent\textbf{Arbitrary-Scale and INR-based SR.} 
Arbitrary-scale SR eliminates the requirement for scale-specific models by treating images as continuous functions. Meta-SR~\cite{hu2019meta} first achieves this via dynamic weight prediction, but INRs such as LIIF~\cite{chen2021liif} have since become the standard paradigm. Subsequent developments, LTE~\cite{lee2022lte} and CiaoSR~\cite{cao2023ciaosr} improve this by incorporating frequency-domain texture estimation. In medical imaging, ArSSR~\cite{wu2022arssr} and SA-INR~\cite{wang2024sainr} adapt INRs to 3D medical volumes, while SAINT~\cite{peng2020saint} uses voxel-spacing-conditioned weights for spatially aware interpolation. Despite their flexibility, coordinate-based MLP inherently suffers from spectral bias~\cite{rahaman2019spectral}, favoring low-frequency structural components over high-frequency details. Furthermore, most existing INR models fail to explicitly incorporate the directional physics of MRI slice acquisition~\cite{pauly1991slr}.

\noindent\textbf{Generative models and adaptive inference.}
Diffusion-based SR methods such as SR3~\cite{saharia2022sr3} and SRDiff~\cite{li2022srdiff} improve perceptual quality but require costly iterative sampling. Acceleration methods such as ResShift~\cite{yue2023resshift} and AddSR~\cite{tai2026addsr} reduce sampling cost through residual shifting or distillation. Rectified Flow~\cite{liu2023flow} and Flow Matching~\cite{lipman2023flow} learn straight-line ODE trajectories, with InstaFlow~\cite{liu2023instaflow} and FlowSR~\cite{xu2025flowsr} enabling few-step generation or restoration. Posterior-mean rectified flow (PMRF)~\cite{ohayon2025pmrf} further formulates image restoration as a two-stage transport process, where a posterior-mean estimate is first obtained and then transported toward the clean image distribution by rectified flow. DRIFT shares this broad two-stage transport view, but specializes it for through-plane MRI SR by introducing acquisition-thickness conditioning, SLR-based slice-profile degradation, metadata-driven adaptive integration, and proximal-thickness trajectory regularization. Unlike image-dependent adaptive inference modules such as AdaDiffSR~\cite{fan2024adadiffsr}, DRIFT uses acquisition metadata to allocate computation without auxiliary difficulty networks, which is suitable for MRI protocols where slice thickness governs through-plane bandwidth loss.

%% file: section/03_method.tex
\section{Method}
\label{sec:method}
We propose DRIFT~(Difficulty-aware RectifIed Flow for Through-plane SR), a two-stage, slice-thickness-conditioned framework for through-plane MRI SR. 
For the $p$-th training patch and a slice-thickness condition indexed by $i$ with input thickness $T_i$, DRIFT reconstructs an isotropic target at thickness $T_{\mathrm{hr}}$ via
\begin{equation}
\mathbf{x}_{p,i} \xrightarrow[\text{Stage 1}]{\text{APN}} \mathbf{z}_{p,i}
\xrightarrow[\text{Stage 2}]{\text{Rectified Flow}} \tilde{\mathbf{y}}_{p,i}.
\label{eq:pipeline}
\end{equation}
Here, $\mathbf{x}_{p,i}\in\mathbb{R}^{H\times W}$ denotes the LR input patch under condition $i$, $\mathbf{y}_{p}\in\mathbb{R}^{H\times W}$ denotes the corresponding HR target patch at $T_{\mathrm{hr}}$, $\mathbf{z}_{p,i}$ is a coarse anatomical estimate, and $\tilde{\mathbf{y}}_{p,i}$ is the final output.
DRIFT is implemented as a 2D slice wise model. Each volume is reconstructed by applying the same thickness-conditioned operator to slices orthogonal to the degraded axis and then composing the predicted slices back into a 3D volume. Therefore, DRIFT does not rely on explicit 3D convolutions or cross-slice communication. Its volumetric continuity instead comes from deterministic APN initialization and shared deterministic rectified flow refinement across neighboring band limited slices.

\subsection{Slice-profile based thick-slice simulation}
\label{sec:simulation}
Through-plane degradation in thick-slice MRI is largely governed by slice-profile induced signal integration along the slice axis.
To approximate clinical slice selection, we synthesize LR inputs of thickness $T_i$ from HR scans at thickness $T_{\mathrm{hr}}$ using an SLR-derived slice-profile model~\cite{pauly1991slr}.

\noindent\textbf{Volume-level simulation.}
Let $\mathbf{Y}\in\mathbb{R}^{H\times W\times D}$ denote an HR 3D volume sampled on an isotropic grid with spacing $T_{\mathrm{hr}}$.
At each iteration, we sample a degradation axis $a\in\{x,y,z\}$ and apply a thickness-specific 1D slice-profile kernel $h_{T_i}$ by convolving $\mathbf{Y}$ only along axis $a$~(leaving the other axes unchanged), yielding a slice-profile integrated volume $\mathbf{Y}'$.
We then resample only along axis $a$ from spacing $T_{\mathrm{hr}}$ to $T_i$~(allowing non-integer factors), and represent the resulting thick-slice volume on the HR grid to form paired training data with $\mathbf{Y}$.
This procedure reproduces the characteristic stair-step morphology observed in reformatted thick-slice MRI.
Implementation details are provided in the supplementary material~(Sec.~S1).

\noindent\textbf{Patch extraction.}
From the simulated thick-slice volume~(represented on the HR grid) and the HR volume $\mathbf{Y}$, we extract 2D slices orthogonal to the degraded axis $a$ and crop aligned patch pairs, corresponding to $\mathbf{x}_{p,i}$ and $\mathbf{y}_p$ defined above.

\subsection{Slice-Thickness Conditioning}
\label{sec:conditioning}

We condition DRIFT on slice-thickness using the inverse thickness $\tau=1/T$, which proxies the effective through-plane bandwidth under slice selection~\cite{pauly1991slr,bernstein2004handbook}.
For condition $i$, we set $\tau_i=1/T_i$ for the input thickness and $\tau_{\mathrm{hr}}=1/T_{\mathrm{hr}}$ for the target thickness. Using $\tau$ reduces the scale disparity between input and target thickness values over clinically common ranges and improves numerical conditioning.
We embed $\tau_i$ and $\tau_{\mathrm{hr}}$ using two lightweight, unshared MLP encoders, concatenate their embeddings, and map the result to a thickness-conditioning vector $\mathbf{c}_i\in\mathbb{R}^{d_c}$~(we use $d_c=512$):
\begin{equation}
\mathbf{c}_i =
\mathrm{MLP}_{\mathrm{cond}}\!\left(
\left[\mathrm{MLP}_{\mathrm{in}}(\tau_i)\;\|\;\mathrm{MLP}_{\mathrm{tgt}}(\tau_{\mathrm{hr}})\right]
\right),
\label{eq:cond}
\end{equation}
where $\|\,$ denotes concatenation, $\mathrm{MLP}_{\mathrm{in}}$ and $\mathrm{MLP}_{\mathrm{tgt}}$ encode the input and target inverse thickness, respectively, and $\mathrm{MLP}_{\mathrm{cond}}$ fuses the two embeddings into the final conditioning vector $\mathbf{c}_i$.
In the main experiments, $T_{\mathrm{hr}}$ is fixed to the native isotropic resolution of each dataset, while the architecture itself accepts both input and target thickness embeddings. Target-continuous operation therefore requires sampling target thicknesses during training, which we examine in the supplementary material.
We inject $\mathbf{c}_i$ into APN residual blocks via Adaptive Group Normalization~(AdaGN)~\cite{dhariwal2021diffusion}:
\begin{equation}
\mathrm{AdaGN}(\mathbf{h},\mathbf{c})=\boldsymbol{\gamma}(\mathbf{c})\odot \mathrm{GN}(\mathbf{h})+\boldsymbol{\beta}(\mathbf{c}),
\label{eq:adagn}
\end{equation}
where $\mathbf{h}\in\mathbb{R}^{C\times H\times W}$ is a feature map, $\mathrm{GN}$ denotes Group Normalization, and $(\boldsymbol{\gamma}(\mathbf{c}),\boldsymbol{\beta}(\mathbf{c}))\in\mathbb{R}^{2\times C}$ are channel-wise scale and shift predicted from $\mathbf{c}$ by a learned linear projection. Architectural details of the MLPs are provided in the supplementary material~(Sec.~S4.1).

\subsection{Stage 1: Structural Manifold Projection}
\label{sec:smp}

Stage~1 is designed to shorten the rectified-flow transport in Stage~2. Given an LR patch $\mathbf{x}_{p,i}$ acquired under thickness condition $i$~(input thickness $T_i$) and the thickness-conditioning vector $\mathbf{c}_i$, the Anatomical Projection Network~(APN) $f_{\phi}$ produces a coarse HR estimate $\mathbf{z}_{p,i}$ at the target thickness $T_{\mathrm{hr}}$:
\begin{equation}
\mathbf{z}_{p,i} = f_{\phi}(\mathbf{x}_{p,i}, \mathbf{c}_i).
\label{eq:apn}
\end{equation}
We refer to this step as structural manifold projection because APN learns a thickness-conditioned mapping that projects samples from the thick-slice input space onto a coarse HR space anchored at $T_{\mathrm{hr}}$. This projection reduces the remaining distance to the target HR distribution, allowing Stage~2 to focus on refining residual high-frequency details with fewer ODE steps.
Beyond reducing the transport distance, APN also provides an image-conditioned initial state for slice-wise generative refinement. Because neighboring thick-slice inputs share anatomical content and are processed by the same deterministic APN, their coarse HR states remain spatially correlated before Stage 2. This differs from slice-wise stochastic generative reconstruction, where independent noise initialization can introduce slice-to-slice variations unless additional noise-correlation heuristics are used.
Through-plane SR is ill-posed, and multiple HR solutions can explain the same thick-slice observation. Under pixel-wise regression, predictors tend to average over plausible solutions, resulting in smooth yet structurally consistent outputs. Accordingly, Stage~1 does not aim to recover the full HR texture distribution; instead, it provides a stable anatomical estimate close to the target HR manifold, which Stage~2 subsequently refines to recover sharp textures and boundaries.
We train APN with a reconstruction objective that combines the Charbonnier penalty~\cite{charbonnier1994two} and SSIM~\cite{wang2004ssim}:
\begin{equation}
\mathcal{L}_{\mathrm{recon}}=
\mathcal{L}_{\mathrm{Char}}(\mathbf{z}_{p,i},\mathbf{y}_{p})
+\lambda_{\mathrm{ssim}}\mathcal{L}_{\mathrm{SSIM}}(\mathbf{z}_{p,i},\mathbf{y}_{p}),
\label{eq:apn_loss}
\end{equation}
where $\lambda_{\mathrm{ssim}}$ weights the SSIM term~(we use $\lambda_{\mathrm{ssim}}=0.5$).

\subsection{Stage 2: Difficulty-Aware Rectified Flow}
\label{sec:darf_train}

Stage~2 refines high-frequency details that are under-recovered by Stage~1 by learning a rectified-flow velocity field~\cite{liu2023flow,lipman2023flow}. During Stage~2 training, we freeze the Stage~1 APN so that the velocity network learns to refine a stable coarse estimate rather than a moving target. The velocity network takes the intermediate image state $\mathbf{s}_{p,i}(t)$ as input, while time $t$ and slice-thickness information modulate intermediate features via AdaGN~(Sec.~\ref{sec:conditioning}).
For training, we sample a normalized time $t\in[0,1]$ and define a straight path between the Stage~1 output $\mathbf{z}_{p,i}$ and the HR target $\mathbf{y}_{p}$:
\begin{equation}
\mathbf{s}_{p,i}(t)=(1-t)\mathbf{z}_{p,i}+t\mathbf{y}_{p},\qquad
\mathbf{u}_{p,i}=\mathbf{y}_{p}-\mathbf{z}_{p,i},
\label{eq:path_u}
\end{equation}
where $\mathbf{u}_{p,i}$ is the constant target velocity along the path. Given $\mathbf{s}_{p,i}(t)$, the velocity network $v_{\theta}$ predicts the rectified-flow velocity, with $t$ and slice-thickness entering only through conditioning. Concretely, we compute a time embedding from a sinusoidal time embedding~\cite{vaswani2017attention}, concatenate it with the thickness-conditioning vector $\mathbf{c}_i$~(Sec.~\ref{sec:conditioning}), and map the concatenation to a joint time--thickness conditioning vector:
\begin{equation}
\mathbf{e}_t=\mathrm{MLP}_{\mathrm{time}}\!\left(\mathrm{SinEmb}(t)\right),\qquad
\mathbf{c}_{t,i}=\mathrm{MLP}_{\mathrm{tt}}\!\left([\mathbf{e}_t\;\|\;\mathbf{c}_i]\right),
\label{eq:ct}
\end{equation}
where $\|\,$ denotes concatenation and $\mathrm{MLP}_{\mathrm{time}}$ and $\mathrm{MLP}_{\mathrm{tt}}$ are lightweight MLPs~(architectural details are provided in the supplementary material Sec.~S4.1). The predicted velocity is
\begin{equation}
\mathbf{v}_{p,i}(t)=v_{\theta}\!\left(\mathbf{s}_{p,i}(t);\mathbf{c}_{t,i}\right),
\label{eq:v_pred}
\end{equation}
where $\mathbf{s}_{p,i}(t)$ is the only network input, and $\mathbf{c}_{t,i}$ is injected into residual blocks via AdaGN.
We train $v_{\theta}$ to match $\mathbf{u}_{p,i}$ using a Huber objective with threshold $\delta=0.1$ and an endpoint-biased timestep distribution $p_{\alpha}(t)$, inspired by the difficulty-aware timestep reweighting principle in RF++~\cite{lee2024rfpp}:
\begin{equation}
\mathcal{L}_{\mathrm{RF}}=
\mathbb{E}_{t\sim p_\alpha(t)}
\left[
\mathrm{Huber}_{\delta}\!\left(\mathbf{v}_{p,i}(t),\mathbf{u}_{p,i}\right)
\right].
\label{eq:rf_loss}
\end{equation}
To sample $t$, we draw $r\sim\mathcal{U}(0,1)$ and apply
\begin{equation}
t=\frac{1}{2}-\frac{1}{2}\,\mathrm{sgn}\!\left(r-\tfrac{1}{2}\right)\,|2r-1|^{1/\alpha},
\label{eq:ushape}
\end{equation}
where $\mathrm{sgn}(\cdot)$ is the sign function and $\alpha$ controls the degree of endpoint emphasis (we use $\alpha=2.0$; see Sec.~S8 in the supplementary material).
Here, $p_\alpha(t)$ denotes the induced distribution of $t$ under Eq.~\ref{eq:ushape}.
This sampling allocates more probability mass to $t\approx 0$ and $t\approx 1$, where refinement is most sensitive in our residual regime. Finally, we zero-initialize the output layer of $v_{\theta}$ so that $\mathbf{v}_{p,i}(t)\approx \mathbf{0}$ at the beginning of training, making Stage~2 initially behave as an identity refinement that preserves $\mathbf{z}_{p,i}$ and then progressively learns residual high-frequency details.

\subsection{CETA: Consistent Endpoint Trajectory Alignment Loss}
\label{sec:ceta}

CETA regularizes thickness-consistent reconstructions for the same underlying anatomy. For a proximal thickness pair $(i,j)$ generated from the same HR patch $\mathbf{y}_{p}$, we set $T_j=T_i-\Delta T$ with $\Delta T=1$\,mm. Here, $\mathbf{z}_{p,k}$ denotes the Stage~1 APN output for condition $k$~(Eq.~\eqref{eq:apn}), for $k\in\{i,j\}$. We form endpoint proxies
\begin{equation}
\tilde{\mathbf{y}}_{p,k}(t)=\mathbf{z}_{p,k}+\mathbf{v}_{p,k}(t),\qquad k\in\{i,j\},
\label{eq:ceta_proxy}
\end{equation}
where $\mathbf{v}_{p,k}(t)$ is the predicted rectified-flow velocity~(Eq.~\eqref{eq:v_pred}). CETA minimizes the squared L2 distance between proxies:
\begin{equation}
\mathcal{L}_{\mathrm{CETA}}=\left\|\tilde{\mathbf{y}}_{p,i}(t)-\tilde{\mathbf{y}}_{p,j}(t)\right\|_2^2.
\label{eq:ceta_loss}
\end{equation}
Using a fixed proximal gap yields a chain of local constraints across the sampled thickness range, which propagates consistency beyond a single pair~(\eg, $T_6 \leftrightarrow T_5 \leftrightarrow \cdots \leftrightarrow T_1$). A very small $\Delta T$ yields nearly identical targets and provides weak alignment signal, while a very large $\Delta T$ forces alignment between substantially different trajectories and can destabilize training. We therefore use $\Delta T=1$\,mm as a practical balance. The Stage~2 objective is $\mathcal{L}_{\mathrm{Stage2}}=\mathcal{L}_{\mathrm{RF}}+\lambda_{\mathrm{ceta}}\mathcal{L}_{\mathrm{CETA}}$~(we use $\lambda_{\mathrm{ceta}}=1.0$. See Sec.~S6 in the supplementary material).
\subsection{PAD and AIS: Physics-Aware Adaptive Inference}
\label{sec:pad_ais}

Inference cost in Stage~2 is dominated by the number of velocity evaluations. Using a fixed step, therefore, wastes computation for thin-slice inputs and can under-refine thick-slice inputs. DRIFT addresses this by selecting the number of Euler steps $N$, equivalently the NFEs of the velocity network, directly from slice-thickness metadata.

\noindent\textbf{Physics-Aware Difficulty~(PAD).}
Given the input thickness $T_i$ and the fixed target thickness $T_{\mathrm{hr}}$~(Sec.~\ref{sec:method}), thicker slices lose a larger fraction of through-plane frequency content due to slice-direction integration. We quantify this protocol difficulty by the normalized bandwidth deficit:
\begin{equation}
\mathrm{PAD}(T_i,T_{\mathrm{hr}})=1-\frac{T_{\mathrm{hr}}}{T_i},
\label{eq:pad}
\end{equation}
where $\mathrm{PAD}\in[0,1)$, equals $0$ when $T_i=T_{\mathrm{hr}}$. PAD serves as a physics-aware degradation severity cue rather than an image difficulty score. It reflects the normalized through-plane bandwidth deficit given by $1 - T_{\mathrm{hr}}/T_i$, which increases as the input slice becomes thicker relative to the target resolution.

\noindent\textbf{Adaptive Integration Scheduler~(AIS).}
AIS maps $\mathrm{PAD}(T_i,T_{\mathrm{hr}})$ to the step budget:
\begin{equation}
N=\mathrm{clamp}\!\Big(\big\lfloor N_{\max}\cdot \mathrm{PAD}(T_i,T_{\mathrm{hr}})\big\rceil,\;N_{\min},\;N_{\max}\Big),
\label{eq:ais}
\end{equation}
where $\lfloor\cdot\rceil$ rounds to the nearest integer and $\mathrm{clamp}(x,a,b)=\min(\max(x,a),b)$. We set $N_{\min}=0$ and $N_{\max}=15$~(See Sec.S7 in the supplementary material). This allocation uses fewer NFEs for easier~(thin-slice) cases and more NFEs for harder~(thick-slice) cases, with negligible overhead since it depends only on slice-thickness metadata.

%% file: section/04_experiments.tex
\section{Experiments}
\label{sec:exp}

\subsection{Experimental Setups}
\label{sec:exp_setup}
\noindent\textbf{Datasets.}
We evaluate on three public brain MRI datasets with isotropic HR volumes as ground truth, namely HCP~\cite{van2013wu}, MIND~\cite{openneuro_ds006391_v2}, and IDEAS~\cite{taylor2025imaging}.
We use T1w and T2w for HCP and MIND, and T1w and FLAIR for IDEAS.
All splits are subject-wise, and we use 890/223, 411/102, and 110/25 subjects for train/test on HCP, MIND, and IDEAS, respectively.
We further perform zero-shot evaluation on real thick-slice MRI without isotropic ground truth using an IRB-approved in-house dataset and the public fastMRI brain dataset~\cite{zbontar2018fastmri}.
The in-house dataset includes T2w and FLAIR scans from 2 healthy volunteers, and the fastMRI subset includes 20 clinically acquired axial T2w volumes.
We additionally use CC359~\cite{souza2018cc359} only for zero-shot PAD scheduling analysis on multi-vendor T1w data.
Additional dataset details are provided in the supplementary material~(Sec.~S2).

\noindent\textbf{Evaluation.}
We evaluate all methods on reconstructed 3D volumes and present peak signal-to-noise ratio~(PSNR) and SSIM~\cite{wang2004ssim}.
For DRIFT and other 2D baselines, we assemble each 3D volume by composing slice-wise predictions and compute PSNR over the reconstructed 3D volume.
SSIM is computed on 2D slices and averaged over axial, coronal, and sagittal directions.
For the IRB-approved in-house dataset, no isotropic ground truth is available, so we present qualitative results only.

\noindent\textbf{Baselines.}
We compare DRIFT against fixed-scale SR methods~\cite{liang2021swinir,song2023afcm,yue2023resshift,lee2023tpdm} and arbitrary-scale SR models~\cite{peng2020saint,chen2021liif,lee2022lte,wang2024sainr,wu2022arssr}.
All baselines are retrained per dataset using official codebases.
Fixed-scale methods train one model per evaluated scale, while arbitrary-scale methods train a single model per dataset.
To enable fair comparison under through-plane SR, we train the 2D restoration and SR models~\cite{liang2021swinir,yue2023resshift} and INR-based baselines~\cite{chen2021liif,lee2022lte,wu2022arssr} using the same SLR-based thick-slice degradation as DRIFT~(Sec.~\ref{sec:simulation}).
This ensures that the LR inputs exhibit the slice axis degradation patterns, including stair-step artifacts in reformatted views.
For baselines whose formulations rely on dedicated specific-plane inputs~\cite{song2023afcm,wang2024sainr}, multi-view inputs~\cite{peng2020saint}, or inverse-problem operators~\cite{lee2023tpdm}, we follow their original input construction and preprocessing.
Baseline-specific details are provided in the supplementary material~(Sec.~S3).

\noindent\textbf{Implementation.}
We implement DRIFT in PyTorch and train on 2D $128\!\times\!128$ patches with a two-stage framework.
Stage~1~(APN) is trained for 100 epochs with a batch size of 64 per GPU and lr $10^{-4}$. In Stage~2, APN is frozen and $v_\theta$ is trained for 100 epochs with a batch size of 25 per GPU and lr $5\!\times\!10^{-5}$.
We use AdamW with cosine annealing and mixed precision on 8$\times$A6000 GPUs.
During training, we synthesize thick-slice inputs using the SLR-based forward model~(Sec.~\ref{sec:simulation}) and sample $T_i \sim \mathcal{U}(T_{\mathrm{hr}}, 6.0]$\,mm.
Training uses random slice and axis sampling, while evaluation uses fixed settings with a fixed seed for reproducibility.
At inference, we apply sliding-window patch inference on each 2D slice with 32-pixel overlap and Gaussian blending, and AIS selects $N\!\in\![0,15]$ NFEs based on PAD~(Sec.~\ref{sec:pad_ais}).
Additional implementation details are provided in the supplementary material~(Sec.~S4).

\begin{figure}[t]
    \centering
    \includegraphics[width=\textwidth]{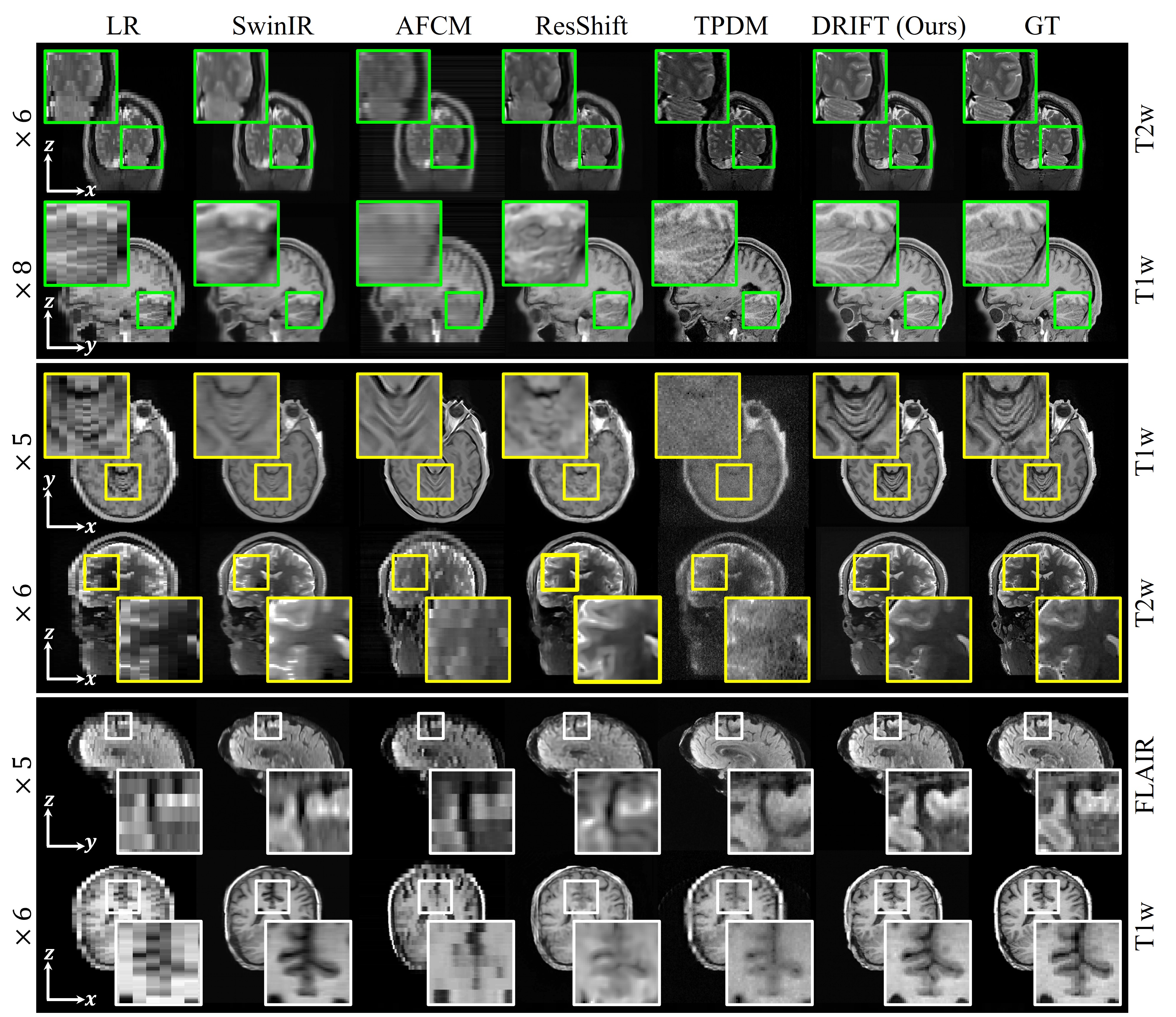}
    \caption{Qualitative comparison with fixed-scale SR baselines on public datasets: HCP~(green), MIND~(yellow), and IDEAS~(white). Results are shown at representative fixed scales within the training range, using HCP at $\times6$~(4.2\,mm) and $\times8$~(5.6\,mm), MIND at $\times5$~(4.5\,mm) and $\times6$~(5.4\,mm), and IDEAS at $\times5$~(5.0\,mm) and $\times6$~(6.0\,mm). Zoom-in boxes highlight regions of interest.}
    \label{fig:qual_all_fixed}
    \vspace{-5.5mm}
\end{figure}
\input{table/main_table}
\subsection{Comparison against SOTA SR baselines on public datasets}
\label{sec:comparison_public}

\noindent\textbf{Qualitative comparisons.}
Fig.~\ref{fig:qual_all_fixed} and Fig.~S1 show visual comparisons on HCP, MIND, and IDEAS.
For fixed-scale SR within the training range, DRIFT best preserves fine anatomical structures and textures while suppressing stair-step artifacts across datasets.
SwinIR~\cite{liang2021swinir} tends to oversmooth details at larger thicknesses.
AFCM~\cite{song2023afcm} exhibits discontinuities along the slice-acquisition direction in non-axial views, which becomes more pronounced at higher scales on HCP and IDEAS.
ResShift~\cite{yue2023resshift} loses fine structures at the largest scale.
TPDM~\cite{lee2023tpdm} reconstructs sharp details but often overestimates patterns, and it can leave residual noise in the output.
The advantage of DRIFT becomes clearer for arbitrary-scale SR under continuous thickness changes, including the out-of-distribution thickness of 6.5\,mm.
Most arbitrary-scale baselines either blur anatomical details or fail to remove stair-step artifacts.
In contrast, DRIFT produces outputs that are most consistent with the ground truth and maintains strong continuity along the slice-acquisition direction.
Compared to 3D network ArSSR~\cite{wu2022arssr}, DRIFT reduces inter-slice inconsistency in challenging regions where stair-step artifacts are prominent~(Fig.~S1 in the supplementary material).
Although DRIFT is slice-wise, its deterministic APN initialization and shared RF refinement reduce visible discontinuities in reformatted views~(Fig.~S8). We further quantify this effect in the supplementary material using through-plane gradient RMSE, where DRIFT shows lower gradient error than TPDM, SA INR, and ArSSR on HCP $\times 8$~(Table.~S5).

\noindent\textbf{Quantitative comparisons.}
Table~\ref{tab:main} indicates PSNR and SSIM on HCP, MIND, and IDEAS under fixed-scale and arbitrary-scale settings, where DRIFT consistently achieves the best performance.
For fixed-scale SR within the training range, DRIFT yields consistent gains across datasets, achieving 37.64/0.952 and 35.93/0.936 on HCP~($\times6$, $\times8$), 32.74/0.904 and 30.02/0.886 on MIND~($\times5$, $\times6$), and 32.75/0.946 and 32.01/0.938 on IDEAS~($\times5$, $\times6$).
Compared to the strongest fixed-scale baseline SwinIR~\cite{liang2021swinir}, DRIFT improves PSNR by up to 3.54\,dB~(HCP), 2.59\,dB~(MIND), and 1.11\,dB~(IDEAS), with SSIM gains up to 0.054.
TPDM~\cite{lee2023tpdm} also appears sensitive to inference hyperparameters and can degrade noticeably across datasets even under the recommended setting of the authors.
For arbitrary-scale SR, DRIFT shows larger advantages since many continuous-scale methods were originally developed under isotropic downsampling.
Across all datasets, DRIFT outperforms the strongest arbitrary-scale baseline, typically LTE~\cite{lee2022lte} or SA-INR~\cite{wang2024sainr}, by 3.29--6.29\,dB in PSNR at 6.0\,mm and 4.71--4.72\,dB at 6.5\,mm, with large SSIM gains.
Notably, DRIFT remains robust at the out-of-distribution thickness of 6.5\,mm, while INR-based baselines such as LIIF~\cite{chen2021liif}, LTE~\cite{lee2022lte}, and ArSSR~\cite{wu2022arssr} show smaller improvements under stair-step artifacts.

\subsection{Zero-Shot Evaluation on Real Thick-Slice MRI}
\label{sec:zeroshot_comparison_real}

We evaluate DRIFT on real thick-slice MRI without isotropic ground truth using an IRB-approved in-house dataset and the public fastMRI brain dataset.
Because full-reference metrics are unavailable, the in-house dataset is used as a prospective qualitative case study across different sequences and acquisition grids, while fastMRI provides a larger public cohort for qualitative comparison and no-reference image-quality evaluation.

\begin{figure}[t]
    \centering
    \includegraphics[width=\textwidth]{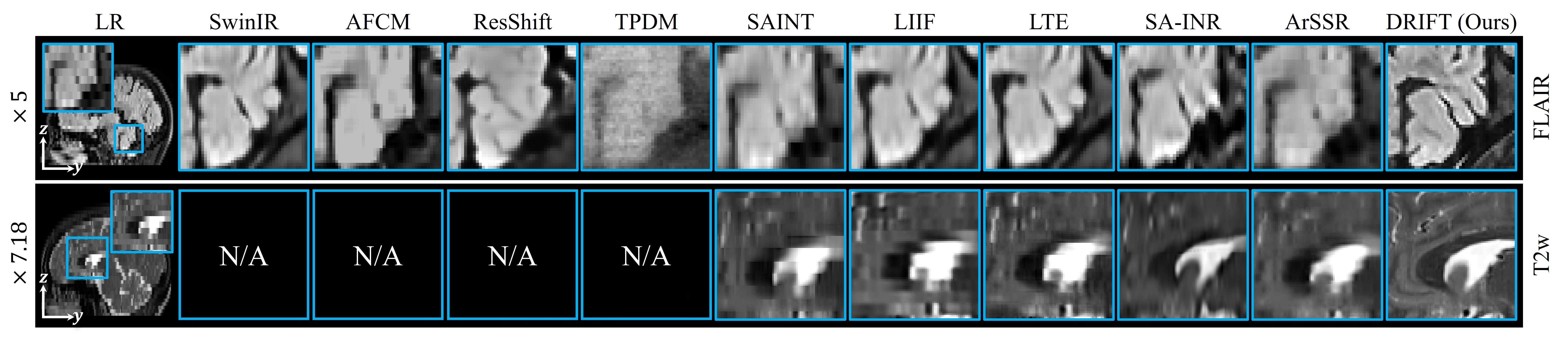}
    \caption{Qualitative comparison on the in-house dataset~(zero-shot).
    IRB-approved thick-slice scans~(5\,mm) for FLAIR and T2w are shown without isotropic ground truth.
    Models are applied without retraining using IDEAS-pretrained weights for FLAIR and HCP-pretrained weights for T2w.
    Zoom-in boxes highlight regions of interest.}
    \label{fig:qual_inhouse}
    \vspace{-3mm}
\end{figure}

For the in-house scans, we use IDEAS-pretrained weights for FLAIR~(1.0\,mm in-plane) and HCP-pretrained weights for T2w~(0.7\,mm in-plane).
As shown in Fig.~\ref{fig:qual_inhouse}, DRIFT reduces stair-step artifacts and preserves anatomical continuity in reformatted views across both sequences.
Fixed-scale baselines cannot be evaluated on the T2w scan because models trained at the required scale are unavailable.
Among arbitrary-scale baselines, several methods leave residual stair-step artifacts, and ArSSR~\cite{wu2022arssr} produces noticeably blurred reconstructions.
TPDM~\cite{lee2023tpdm} shows weaker cross-dataset generalization and fails to recover coherent structures on the in-house scans.

For fastMRI, we apply the HCP-pretrained DRIFT model to clinically acquired T2w scans with 5\,mm slice thickness and compare against foundation model BME-X~\cite{sun2025bmex}.
Qualitative comparisons show that DRIFT produces sharper coronal reformats with fewer through-plane artifacts than BME-X~\cite{sun2025bmex}.
DRIFT also achieves higher sharpness (0.291 vs. 0.238) and lower NIQE and BRISQUE (5.30 and 21.16 vs. 7.22 and 62.46), supporting its zero-shot use on real thick-slice MRI without retraining or dataset-specific tuning~(Fig.~S9 and Table~S6 in the supplementary material).

\subsection{Efficiency--Fidelity Trade-off}
\label{sec:efficiency}

In this section, we analyze an efficiency--fidelity trade-off between reconstruction fidelity and inference cost from two complementary perspectives and further report standardized model cost in the supplementary material.

\begin{figure}[t]
    \centering
    \begin{minipage}[t]{0.475\columnwidth}
        \centering
        \includegraphics[width=0.98\linewidth]{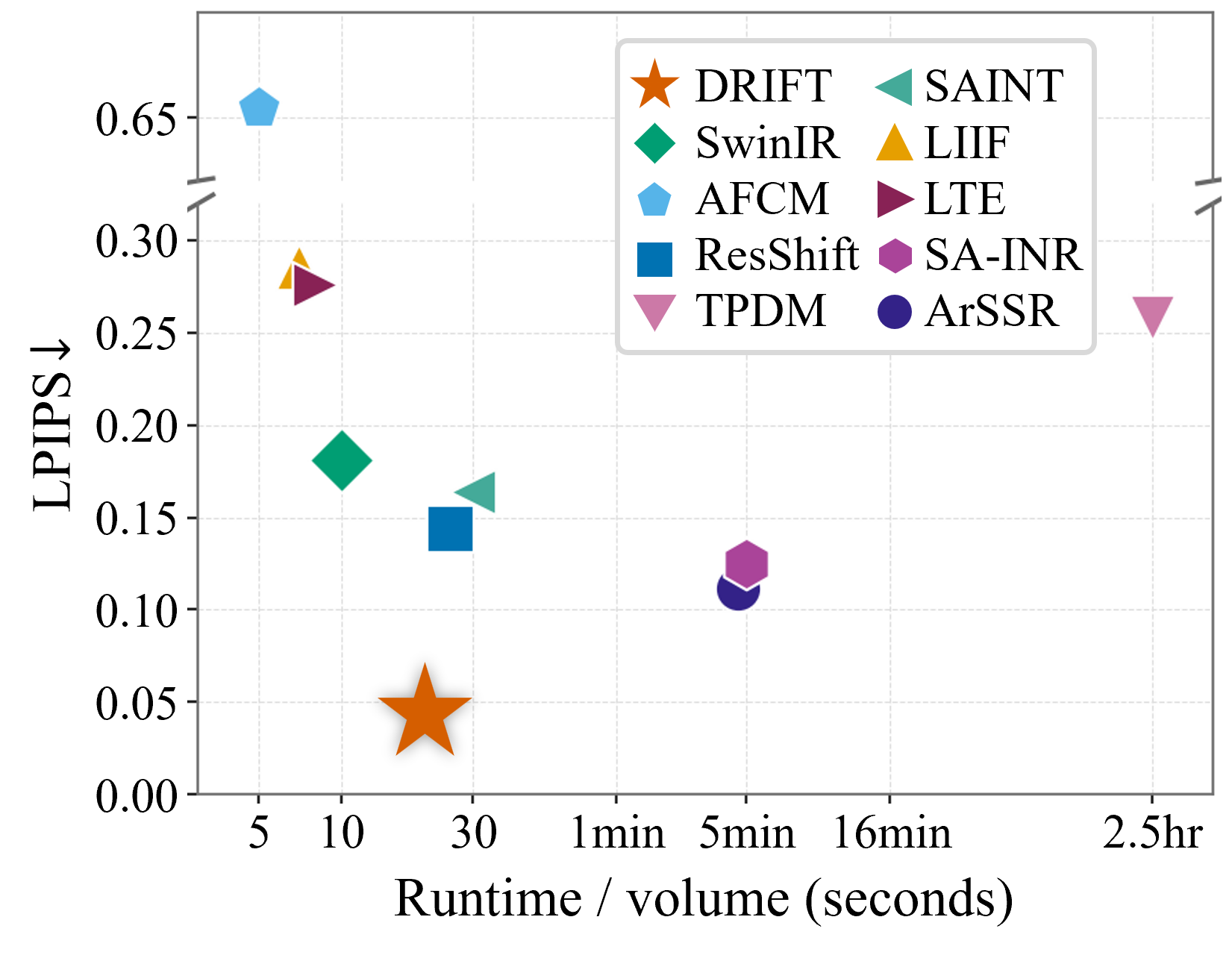}
        \vspace{-4mm}
        \caption{Perceptual efficiency-fidelity trade-off on HCP $\times$8~(5.6\,mm). Per-volume runtime versus LPIPS.}
        \label{fig:perceptual_tradeoff}
    \end{minipage}
    \hfill
    \begin{minipage}[t]{0.50\columnwidth}
        \centering
        \includegraphics[width=0.98\linewidth]{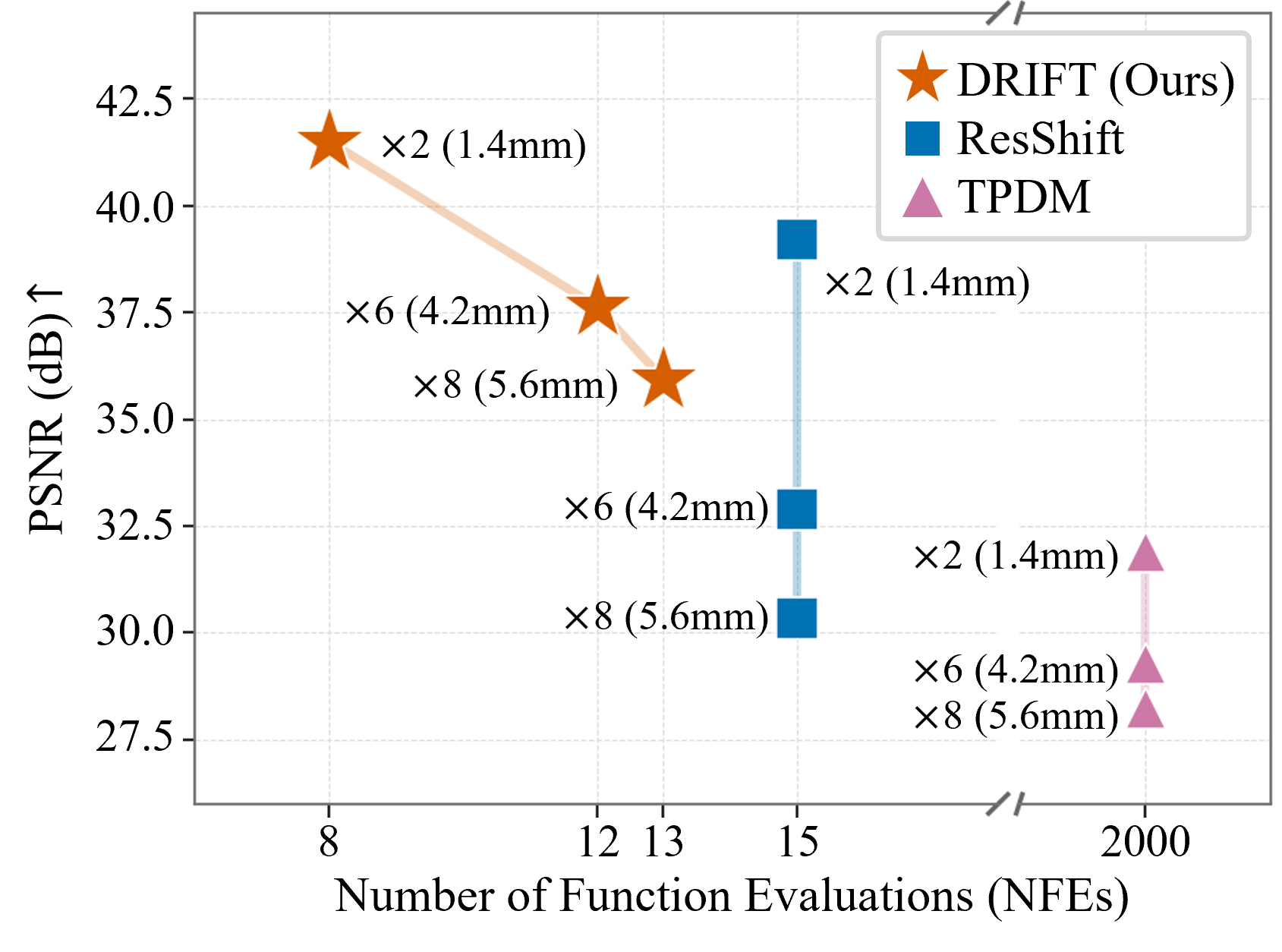}
        \vspace{-4mm}
        \caption{Distortion efficiency-fidelity trade-off against multi-step diffusion baselines on HCP. PSNR versus NFEs at $\times2$~(1.4\,mm), $\times6$~(4.2\,mm), and $\times8$~(5.6\,mm).}
        \label{fig:distortion_tradeoff}
    \end{minipage}
\end{figure}

\noindent\textbf{Perceptual efficiency--fidelity~(runtime vs. LPIPS).}
Fig.~\ref{fig:perceptual_tradeoff} indicates per-volume runtime against LPIPS on HCP $\times$8~(5.6\,mm$\rightarrow$0.7\,mm).
Most feed-forward baselines~\cite{liang2021swinir,song2023afcm,chen2021liif,lee2022lte} are fast but show higher LPIPS due to oversmoothing.
Several volumetric or coordinate-based feed-forward models, including SAINT~\cite{peng2020saint}, SA-INR~\cite{wang2024sainr}, and ArSSR~\cite{wu2022arssr}, can be slower than DRIFT in this setting.
DRIFT achieves the lowest LPIPS~(0.043) at about 20\,s per volume, improving over ArSSR~\cite{wu2022arssr}~(0.112) by $\times$3.3 while being $\times$14 faster.
TPDM~\cite{lee2023tpdm} is about $\times$450 slower yet yields $\times$6 worse LPIPS than DRIFT.

\noindent\textbf{Distortion efficiency--fidelity~(NFEs vs. PSNR).}
Fig.~\ref{fig:distortion_tradeoff} shows PSNR versus NFEs against multi-step diffusion baselines on HCP.
ResShift~\cite{yue2023resshift} runs with a small fixed step count~(15 NFEs), which does not adapt to thickness-dependent difficulty.
TPDM~\cite{lee2023tpdm} requires 2{,}000 steps while producing lower PSNR.
In contrast, DRIFT achieves higher PSNR with fewer evaluations, using 8, 12, and 13 NFEs at $\times2$, $\times6$, and $\times8$, respectively.
Overall, DRIFT is slice-thickness-aware and adaptively allocates NFEs, achieving higher PSNR than fixed-step diffusion baselines while using fewer function evaluations.

\noindent\textbf{Standardized model cost.}
We further report standardized cost and compact variant results in the supplementary material~(Sec.~S4.2).
A smaller DRIFT-Small variant keeps the same two-stage design and adaptive integration mechanism while reducing parameters, FLOPs, and peak VRAM by $\times$3.3, $\times$4.0, and $\times$5.3 relative to DRIFT-Large.
Its performance remains competitive across HCP, MIND, and IDEAS, indicating that the efficiency and fidelity behavior of DRIFT is not explained by model size alone.

\subsection{Ablation Study}
\label{sec:ablation_study}

\noindent\textbf{PAD versus Image-Dependent Difficulty.}
On multi-vendor CC359~\cite{souza2018cc359}, an AdaDiffSR-style image score shows negligible correlation with HF-PSNR within fixed input-thickness groups.
Image-only and hybrid scheduling also underperform PAD, supporting slice-thickness-based scheduling in this controlled through-plane SR setting~(Sec.~S11).

\noindent\textbf{Ablation on the Two-Stage Design.}
\label{sec:ablation_two_stage}
We compare Stage~1 only, Stage~2 only, and full DRIFT.
Stage~1 removes coarse stair-step artifacts but oversmooths fine detail, while Stage~2 without APN improves local sharpness but lacks stable anatomical initialization at larger slice thicknesses.
Full DRIFT combines anatomical initialization with residual rectified-flow refinement and gives the best quantitative and qualitative results~(Fig.~S2).


\input{table/ablation_table}

\noindent\textbf{Ablation on Each Stage~2 Component.}
\label{sec:ablation_stage2_component}
Table~\ref{tab:ablation} quantifies the contribution of each Stage~2 component on HCP.
Removing slice-thickness conditioning causes the largest degradation, showing that acquisition conditioning is essential for heterogeneous thick-slice inputs.
CETA and U-shaped timestep sampling further improve the result, indicating complementary roles of trajectory regularization and timestep prioritization.

\noindent\textbf{Ablation on the CETA Proximal Gap.}
\label{sec:ablation_ceta_gap}
Table~\ref{tab:ceta_gap} evaluates the gap $\Delta T$ used to form CETA loss pairs.
Random CETA pairs improve over the no CETA setting in Table~\ref{tab:ablation}, but remain below the fixed 1.0\,mm gap despite moderate coverage, indicating that pair coverage alone is not sufficient.
Increasing $\Delta T$ can provide a stronger cross-thickness constraint, but it also reduces coverage within the training range and aligns more dissimilar conditions, which may make optimization less stable.
Conversely, a small gap such as 0.5\,mm yields high coverage but provides a weaker learning signal because paired conditions become nearly indistinguishable, making the alignment close to an identity constraint.
Across random and fixed-gap settings, $\Delta T=1.0$\,mm provides the best trade-off between constraint strength, locality, and coverage, achieving the best PSNR, SSIM, and LPIPS.

\noindent\textbf{Additional evaluations.}
Additional downstream segmentation results and multi-planar volumetric visualizations are provided in the supplementary material to assess anatomical utility and volumetric reformat quality.~(Sec.~S14 and Sec.~S15).

%% file: table/main_table.tex
\newcommand{\ps}[2]{#1/#2} 
\newcommand{\best}[1]{\textcolor{eccvblue}{\textbf{#1}}} 

\begin{table}[t]
\centering
\caption{Quantitative comparison with state-of-the-art methods on HCP, MIND, and IDEAS for through-plane MRI SR. We report PSNR$\uparrow$/SSIM$\uparrow$ for fixed-scale and arbitrary-scale settings. The best results are highlighted in \textcolor{eccvblue}{blue}. Additional scales are in the supplementary.}
\label{tab:main}
\scriptsize
\setlength{\tabcolsep}{2.0pt}
\renewcommand{\arraystretch}{0.9}
\begin{adjustbox}{width=\columnwidth}
\begin{tabular}{l cc cc cc}
\toprule
Method &
\multicolumn{2}{c}{HCP} &
\multicolumn{2}{c}{MIND} &
\multicolumn{2}{c}{IDEAS} \\
\cmidrule(lr){2-3}\cmidrule(lr){4-5}\cmidrule(lr){6-7}

\textbf{Fixed-scale} &
\makecell{\rule{0pt}{2.6ex}$\times6$\\(4.2mm)} &
\makecell{\rule{0pt}{2.6ex}$\times8$\\(5.6mm)} &
\makecell{\rule{0pt}{2.6ex}$\times5$\\(4.5mm)} &
\makecell{\rule{0pt}{2.6ex}$\times6$\\(5.4mm)} &
\makecell{\rule{0pt}{2.6ex}$\times5$\\(5.0mm)} &
\makecell{\rule{0pt}{2.6ex}$\times6$\\(6.0mm)} \\
\midrule
SwinIR~\cite{liang2021swinir}     & \ps{34.10}{0.901} & \ps{33.13}{0.882} & \ps{30.15}{0.874} & \ps{29.01}{0.838} & \ps{31.64}{0.924} & \ps{31.62}{0.914} \\
AFCM~\cite{song2023afcm}         & \ps{27.11}{0.642} & \ps{26.08}{0.501} & \ps{27.74}{0.728} & \ps{26.02}{0.713} & \ps{23.89}{0.825} & \ps{22.55}{0.796} \\
ResShift~\cite{yue2023resshift}  & \ps{32.89}{0.888} & \ps{30.34}{0.853} & \ps{29.48}{0.839} & \ps{29.00}{0.833} & \ps{31.93}{0.910} & \ps{30.40}{0.899} \\
TPDM~\cite{lee2023tpdm}          & \ps{29.31}{0.798} & \ps{28.24}{0.755} & \ps{22.50}{0.452} & \ps{22.11}{0.413} & \ps{26.64}{0.744} & \ps{26.11}{0.720} \\
DRIFT (Ours)                     & \best{\ps{37.64}{0.952}} & \best{\ps{35.93}{0.936}} & \best{\ps{32.74}{0.904}} & \best{\ps{30.02}{0.886}} & \best{\ps{32.75}{0.946}} & \best{\ps{32.01}{0.938}} \\
\midrule

\textbf{Arbitrary-scale} &
\makecell{\rule{0pt}{2.6ex}$\times8.57$\\(6.0mm)} &
\makecell{\rule{0pt}{2.6ex}$\times9.28$\\(6.5mm)} &
\makecell{\rule{0pt}{2.6ex}$\times6.67$\\(6.0mm)} &
\makecell{\rule{0pt}{2.6ex}$\times7.22$\\(6.5mm)} &
\makecell{\rule{0pt}{2.6ex}$\times5.50$\\(5.5mm)} &
\makecell{\rule{0pt}{2.6ex}$\times6.50$\\(6.5mm)} \\
\midrule
SAINT~\cite{peng2020saint}       & \ps{25.09}{0.686} & \ps{24.77}{0.666} & \ps{23.39}{0.721} & \ps{23.05}{0.701} & \ps{24.63}{0.649} & \ps{24.33}{0.627} \\
LIIF~\cite{chen2021liif}         & \ps{26.60}{0.801} & \ps{26.53}{0.782} & \ps{25.27}{0.774} & \ps{25.11}{0.773} & \ps{27.24}{0.850} & \ps{26.98}{0.842} \\
LTE~\cite{lee2022lte}            & \ps{26.68}{0.803} & \ps{26.65}{0.802} & \ps{25.24}{0.772} & \ps{25.20}{0.771} & \ps{27.35}{0.841} & \ps{27.27}{0.830} \\
SA-INR~\cite{wang2024sainr}       & \ps{25.86}{0.694} & \ps{25.18}{0.665} & \ps{26.22}{0.810} & \ps{25.12}{0.774} & \ps{26.18}{0.745} & \ps{26.10}{0.735} \\
ArSSR~\cite{wu2022arssr}          & \ps{25.15}{0.699} & \ps{25.10}{0.685} & \ps{24.40}{0.758} & \ps{24.00}{0.741} & \ps{24.85}{0.682} & \ps{24.50}{0.679} \\
DRIFT (Ours)                      & \best{\ps{32.97}{0.891}} & \best{\ps{32.90}{0.880}} & \best{\ps{29.51}{0.882}} & \best{\ps{29.50}{0.881}} & \best{\ps{32.05}{0.945}} & \best{\ps{31.99}{0.935}} \\
\bottomrule
\end{tabular}
\end{adjustbox}
\vspace{-2mm}
\end{table}

%% file: table/ablation_table.tex
\begin{table*}[t]
  \centering

  \begin{minipage}[t]{0.50\textwidth}
    \centering
    \small
    \captionof{table}{Stage~2 component ablation on HCP. We remove one component while keeping the Stage~1 APN fixed. Metrics are averaged over $T_{\mathrm{i}}\in\{1.5,3.0,5.0\}$\,mm with $T_{\mathrm{hr}}=0.7$\,mm. CETA is N/A without thickness conditioning.}
    \label{tab:ablation}
    \vspace{-5pt}
    \setlength{\tabcolsep}{4pt}
    \renewcommand{\arraystretch}{0.92}

    \resizebox{\linewidth}{!}{%
    \begin{tabular}{@{}ccc|ccc@{}}
      \toprule
      \makecell{Thickness\\Cond.} &
      \makecell{U-shaped\\Sampling} &
      \makecell{CETA loss\\($\lambda_{\mathrm{CETA}}=1$)} &
      PSNR$\uparrow$ & SSIM$\uparrow$ & LPIPS$\downarrow$ \\
      \midrule
      \cmark & \cmark & \xmark & 36.92 & 0.9416 & 0.0308 \\
      \xmark & \cmark & N/A    & 34.82 & 0.8986 & 0.0491 \\
      \cmark & \xmark & \cmark & 38.12 & 0.9458 & 0.0342 \\
      \cmark & \cmark & \cmark & \best{40.85} & \best{0.9634} & \best{0.0229} \\
      \bottomrule
    \end{tabular}%
    }
  \end{minipage}\hfill
  \begin{minipage}[t]{0.46\textwidth}
    \centering
    \small
   \captionof{table}{CETA gap ablation on HCP. Random denotes naive random CETA pairs, and Coverage denotes the supervised fraction of the thickness range. Metrics are averaged over $T_{\mathrm{i}}\in\{1.5,3.0,5.0\}$\,mm with $T_{\mathrm{hr}}=0.7$\,mm.}
    \label{tab:ceta_gap}
    \vspace{-5pt}
    \setlength{\tabcolsep}{4pt}
    \renewcommand{\arraystretch}{0.75}

    \resizebox{\linewidth}{!}{%
    \begin{tabular}{@{}ccccc@{}}
    \toprule
    Gap $\Delta T$ & Coverage & PSNR$\uparrow$ & SSIM$\uparrow$ & LPIPS$\downarrow$ \\
    \midrule
    Random & 65\% & 38.23 & 0.9460 & 0.0290 \\
    0.5\,mm & 91\% & 37.24 & 0.9430 & 0.0301 \\
    \best{1.0\,mm} & \best{81\%} & \best{40.85} & \best{0.9634} & \best{0.0229} \\
    2.0\,mm & 62\% & 39.10 & 0.9540 & 0.0265 \\
    4.0\,mm & 25\% & 37.58 & 0.9445 & 0.0338 \\
    \bottomrule
    \end{tabular}%
    }
  \end{minipage}
\end{table*}

%% file: section/05_conclusion.tex
\section{Conclusion}
\label{sec:conclusion}
We propose DRIFT, a two-stage difficulty-aware rectified flow framework for through-plane MRI SR, with thickness conditioning for continuous input slice thicknesses.
APN projects thick-slice inputs onto a coarse HR manifold, reducing the residual transport solved by Stage~2.
PAD-guided AIS allocates ODE steps according to slice-thickness difficulty, while CETA regularizes proximal thickness trajectories for consistent reconstructions.
Across public datasets and real thick-slice MRI, DRIFT improves fidelity and sampling efficiency over regression, INR, and diffusion baselines.
A current limitation is that the main experiments use fixed native target thicknesses.
Extending target-thickness sampling across resolutions and anatomies remains an important direction for cross-protocol generalization.

%% file: supple.tex


\setcounter{section}{0}
\setcounter{subsection}{0}
\setcounter{figure}{0}
\setcounter{table}{0}
\setcounter{equation}{0}

\renewcommand{\thesection}{S\arabic{section}}
\renewcommand{\thesubsection}{S\arabic{section}.\arabic{subsection}}
\renewcommand{\theHsection}{S\arabic{section}}
\renewcommand{\theHsubsection}{S\arabic{section}.\arabic{subsection}}
\renewcommand{\thefigure}{S\arabic{figure}}
\renewcommand{\thetable}{S\arabic{table}}
\renewcommand{\theequation}{S\arabic{equation}}

\renewcommand{\theHfigure}{S\arabic{figure}}
\renewcommand{\theHtable}{S\arabic{table}}
\renewcommand{\theHequation}{S\arabic{equation}}

\providecommand{\figureautorefname}{Fig.}
\providecommand{\tableautorefname}{Tab.}
\providecommand{\sectionautorefname}{Sec.}
\providecommand{\subsectionautorefname}{Sec.}
\providecommand{\equationautorefname}{Eq.}

\input{supp/s1_simulation}
\input{supp/s2_dataset_detail}
\input{supp/s3_baseline_detail}
\input{supp/s4_implementation_detail}
\input{supp/fig_qualitative_arbitrary_sr}
\input{supp/s5_ablation_stage_quant}
\input{supp/s6_ceta_weight}

\input{supp/s7_optimal_nfe_max}
\input{supp/s8_ablation_endpoint_bias_timestep_sampling}
\input{supp/s9_ablation_ais}
\input{supp/s10_ablation_pad_justification}

\input{supp/s11_pad_vs_image_difficulty}
\input{supp/s12_ablation_ceta_timestep_sensitivity}
\input{supp/s13_ablation_continuity}
\input{supp/s14_downstream_segmentation}

\input{supp/s15_multiplanar_vol_visualization}

%% file: supp/s1_simulation.tex
\section{Slice-profile thick-slice simulation: implementation details}
\label{sec:supp_simulation}
\setcounter{equation}{0}

\noindent\textbf{Notation.}
$T_{\mathrm{hr}}$ denotes the HR target slice-thickness and $T_i$ the sampled input thickness.
Let $\mathbf{Y}$ be an HR 3D volume at spacing $T_{\mathrm{hr}}$.
We randomly choose a degradation axis $a\in\{x,y,z\}$ and use an SLR-derived slice profile kernel $h_{T_i}$ for thickness $T_i$~\cite{pauly1991slr,bernstein2004handbook}.
Operators $*_{a}$, $\mathcal{R}^{(a)}_{T_i \leftarrow T_{\mathrm{hr}}}(\cdot)$, and $\mathcal{I}^{(a)}_{\mathrm{NN}}(\cdot)$ denote axis-only convolution, axis-only resampling, and axis-only nearest-neighbor interpolation, respectively.

\subsubsection{Slice-profile kernel.}
For each $T_i$, we construct a discrete 1D kernel $h_{T_i}$ on the HR grid (spacing $T_{\mathrm{hr}}$) from the corresponding SLR slice profile.
We normalize it to preserve DC gain,
$\sum_n h_{T_i}[n]=1$,
and apply symmetric padding along axis $a$ during convolution to avoid boundary attenuation.

\subsubsection{Thick-slice formation (axis-only).}
We simulate through-plane signal integration by convolving only along the selected axis:
\begin{equation}
\mathbf{Y}' = \mathbf{Y} *_{a} h_{T_i}.
\label{eq:supp_slr_conv}
\end{equation}
We then resample only along axis $a$ from spacing $T_{\mathrm{hr}}$ to $T_i$ using center-aligned linear resampling (non-integer factors are allowed):
\begin{equation}
\mathbf{X}^{(a)}_{i,\downarrow} =
\mathcal{R}^{(a)}_{T_i \leftarrow T_{\mathrm{hr}}}\!\left(\mathbf{Y}'\right).
\label{eq:supp_resample_down}
\end{equation}
For paired training on the HR grid while preserving stair-step morphology, we map the thick-slice volume back to the HR grid using nearest-neighbor interpolation along the same axis:
\begin{equation}
\mathbf{X}^{(a)}_{i} =
\mathcal{I}^{(a)}_{\mathrm{NN}}\!\left(\mathbf{X}^{(a)}_{i,\downarrow}\right).
\label{eq:supp_nn_up}
\end{equation}

\subsubsection{Multi-plane 2D sampling (training pairs).}
From $(\mathbf{X}^{(a)}_{i},\mathbf{Y})$, we sample 2D slices from planes orthogonal to $a$ so that stair-step artifacts are explicit in the inputs.
For each subject, we randomly select $K$ slice indices per orthogonal axis (we use $K=100$) and extract aligned patch pairs $(\mathbf{x}_{p,i},\mathbf{y}_p)$.
Axis randomization across iterations improves robustness to different reformatting views.

%% file: supp/s2_dataset_detail.tex
\section{Dataset details}
\label{sec:supp_datasets}

\noindent\textbf{Public datasets.}
HCP~\cite{van2013wu} includes 1,113 healthy subjects at 0.7\,mm isotropic (T1w, T2w), and we use 890/223 for train/test with volumes of $320\times320\times320$.
MIND~\cite{openneuro_ds006391_v2} includes 513 subjects at 0.9\,mm isotropic (T1w, T2w), and we use 411/102 with volumes of $256\times256\times256$.
IDEAS~\cite{taylor2025imaging} includes 135 subjects at 1.0\,mm isotropic (T1w, FLAIR), and we use 110/25 with volumes of $256\times256\times256$.

\noindent\textbf{Slice sampling.}
For training, we sample 2D slices from axial, coronal, and sagittal planes and extract $128\times128$ patches.
We sample 100 slices per plane for each subject and modality, yielding 300 slices per subject and modality.

\noindent\textbf{In-house dataset.}
We use an IRB-approved in-house dataset of 2 healthy volunteers scanned on a 3\,T MAGNETOM Vida (Siemens Healthineers).
We acquire thick-slice FLAIR with 1.0\,mm in-plane resolution and thick-slice T2w with 0.7\,mm in-plane resolution, both at 5\,mm slice-thickness.
The acquisition grids are $368\times368\times52$ for T2w and $256\times256\times52$ for FLAIR to match through-plane coverage to the in-plane field of view.
No isotropic ground truth is available, so we report qualitative results only.

\noindent\textbf{fastMRI dataset.}
We use a subset of the public fastMRI brain dataset~\cite{zbontar2018fastmri}, which contains clinically acquired axial brain MRI scans from multiple sites and Siemens 1.5\,T and 3\,T scanners.
We select 20 axial T2w volumes with $0.69\times0.69\times5\,\mathrm{mm}^3$ resolution and 16 slices per volume.
The selected scans were acquired on Siemens Avanto and Aera 1.5\,T scanners for 17 volumes and a Siemens Biograph 3\,T scanner for 3 volumes.
No isotropic ground truth is available, so we report qualitative comparisons and no-reference image-quality metrics.

\noindent\textbf{CC359 dataset.}
We use CC359~\cite{souza2018cc359} for zero-shot analysis of PAD scheduling under multi-vendor and multi-field-strength variation.
CC359 contains T1w brain MRI acquisitions from six scanner models across GE, Philips, and Siemens scanners at 1.5\,T and 3\,T.
We use the 235 volumes provided at 1.0\,mm isotropic resolution as high-resolution ground truth. 
To obtain paired low-resolution inputs, we apply the same slice-profile based degradation used in the main experiments along the through-plane axis.
All 235 volumes are used only for evaluation, and no CC359 data are used for training or fine-tuning.

%% file: supp/s3_baseline_detail.tex
\section{Baseline-specific configurations}
\label{sec:supp_baselines}

\noindent\textbf{AFCM}~\cite{song2023afcm}.
We follow the released training protocol of AFCM, which requires pre-resampling the LR input to the target shape using nearest-neighbor interpolation and enforcing strict slice alignment across modalities for each subject.
This preprocessing step is part of the alias-aware design of AFCM for cross-modality synthesis and SR.
For SR, we adopt the official implementation and configuration of AFCM.

\noindent\textbf{TPDM}~\cite{lee2023tpdm}.
TPDM solves through-plane SR as a 3D inverse problem using two perpendicular 2D diffusion priors.
In their MR-ZSR setting, the measurement is formed by merging adjacent slices along the slice axis and averaging within each group, and the posterior sampling step uses a slightly modified averaging kernel during enforcement of measurement consistency.
We follow their original formulation for data construction and reconstruction.

\noindent\textbf{SAINT}~\cite{peng2020saint}.
SAINT performs slice synthesis by exploiting multi-view consistency.
It first estimates missing slices from sagittal and coronal views, then refines axial slices using a residual fusion network to reduce directional artifacts.
Voxel spacing information is used as an explicit input to guide interpolation strength.
We follow the original SAINT pipeline for input formation and reconstruction.

\noindent\textbf{MRI INR-based baselines.}
SA-INR~\cite{wang2024sainr} and ArSSR~\cite{wu2022arssr} are INR-based methods designed for 3D MR reconstruction.
SA-INR represents an MR volume as a continuous implicit function of 3D coordinates and reconstructs arbitrary slice spacings by resampling coordinates, with spatial attention to improve local consistency.
ArSSR learns an implicit voxel function from paired LR and HR examples for arbitrary-scale 3D MR reconstruction.
For both methods, we train and evaluate them in their original 3D coordinate-based setting, using LR-HR pairs generated with the same SLR-based through-plane degradation as DRIFT for fair comparison.

\noindent\textbf{Natural-image INR baselines.}
LIIF~\cite{chen2021liif} and LTE~\cite{lee2022lte} are originally developed for isotropic natural-image downsampling.
To expose them to through-plane artifacts, we generate LR inputs using the same SLR-based thick-slice degradation as DRIFT and provide anisotropic scale information through coordinate conditioning.

\noindent\textbf{2D restoration baseline.}
SwinIR~\cite{liang2021swinir} is a 2D image restoration model rather than an INR-based method.
We disable the upsampling module by setting the upsampling factor to $\times1$, and use it as a same-resolution restoration network on the degraded inputs.

%% file: supp/s4_implementation_detail.tex
\newcommand{\ps}[2]{#1/#2} 
\newcommand{\best}[1]{\textcolor{eccvblue}{\textbf{#1}}} 

\section{Implementation Details}
\label{sec:supp_implement}
\subsection{Network Architecture}
\label{sec:supp_network_architecture}
Both the APN (Stage~1) and velocity network $v_\theta$ (Stage~2) use
a 2D U-Net with four encoder levels and a symmetric decoder.
We use the \texttt{large} configuration: base channel dimension $C{=}96$
with channel multipliers $(1, 2, 4, 8)$, yielding feature maps of
$[96, 192, 384, 768]$ channels at each resolution level.
Each encoder level contains 2 residual blocks, and each decoder level
contains 3 residual blocks (one extra to accommodate the skip connection).
The bottleneck consists of 2 residual blocks at 768 channels.
Downsampling is performed via strided convolution ($3{\times}3$, stride 2),
and upsampling via bilinear interpolation followed by a $3{\times}3$
convolution. Skip connections use channel-wise concatenation.

Each residual block consists of two $3{\times}3$ convolutions with
SiLU activation and Adaptive Group Normalization (AdaGN, 32 groups),
where the affine parameters (scale and shift) are predicted from
the conditioning vector.
In Stage~1, the conditioning vector (512-d) is derived solely from the
acquisition protocol: each thickness $T$ is converted to
$\tau{=}1/T$ and independently projected through a 2-layer MLP
($1 \to 128 \to 256$), then concatenated and fused.
In Stage~2, an additional sinusoidal positional embedding encodes
the flow timestep $t \in [0,1]$ into a 256-d vector, which is
concatenated with the 512-d protocol embedding and projected
to a 768-d conditioning vector.
The output convolution of Stage~2 is zero-initialized so that the
initial velocity prediction $v_\theta \approx 0$ and training starts
near the Stage~1 output.
Stage~1 has 127.9M parameters and Stage~2 has 139.0M parameters, yielding 266.9M parameters for DRIFT-Large.

\subsection{Standardized Model Cost and Compact Variant}
\label{sec:supp_standardized_cost}
To complement the runtime and NFE analysis in Sec.~4.4, we profile DRIFT variants and representative baselines under a standardized setting.
We report parameters, FLOPs, and peak VRAM using the same input size, batch size, and hardware.
FLOPs are measured on $128\times128$ inputs, and peak VRAM is measured with batch size 1 on the same GPU.
Total sampling cost further depends on the number of function evaluations, which is analyzed in Sec.~4.4.

\begin{table}[t]
\centering
\caption{Standardized model cost across baselines and DRIFT variants.}
\label{tab:supp_standardized_cost}
\setlength{\tabcolsep}{6pt}
\begin{tabular}{lccc}
\toprule
Method & Params (M) & FLOPs (G) & Peak VRAM (MB) \\
\midrule
DRIFT-Large & 266.9 & 173.0 & 1147 \\
DRIFT-Small & \best{80.9} & \best{43.5} & \best{218} \\
ArSSR~\cite{wu2022arssr} & 6.3 & 196.8 & 202 \\
ResShift~\cite{yue2023resshift} & 118.3 & 75.5 & 611 \\
TPDM~\cite{lee2023tpdm} & 44.1 & 34.1 & 222 \\
SAINT~\cite{peng2020saint} & 4.5 & 72.4 & 167 \\
SwinIR~\cite{liang2021swinir} & 3.1 & 55.5 & 146 \\
LIIF~\cite{chen2021liif} & 1.5 & 23.5 & 133 \\
LTE~\cite{lee2022lte} & 1.7 & 19.3 & 139 \\
SA-INR~\cite{wang2024sainr} & 3.0 & 96.8 & 1052 \\
\bottomrule
\end{tabular}
\end{table}

Table~\ref{tab:supp_standardized_cost} shows that DRIFT-Small reduces parameters, FLOPs, and peak VRAM by $\times$3.3, $\times$4.0, and $\times$5.3 relative to DRIFT-Large, respectively.
Although reducing capacity lowers PSNR, DRIFT-Small retains the same projection to rectified flow design and the same adaptive integration mechanism.
It achieves PSNR values of 32.65, 29.47, and 31.04 dB on HCP, MIND, and IDEAS, respectively, compared with 35.93, 30.02, and 32.01 dB for DRIFT-Large.
The larger performance gap on HCP is consistent with its finer 0.7\,mm target resolution, which places a stronger demand on high-frequency detail recovery.
On MIND and IDEAS, which use coarser 0.9\,mm and 1.0\,mm target resolutions, DRIFT-Small remains close to DRIFT-Large while substantially reducing model cost.
This compact variant indicates that the efficiency and fidelity behavior of DRIFT is not explained by model size alone.

\subsection{Data Preprocessing}
\label{sec:supp_data_preprocessing}
\noindent\textbf{Intensity normalization.}
Each 2D slice is independently min-max normalized to $[0,1]$ and then
linearly mapped to $[-1,1]$.

\noindent\textbf{Sampling.}
During training, $T_\text{i}$ is uniformly sampled from
$(T_\text{native}, 6.0]$\,mm and $T_\text{hr}$ is fixed at
$T_\text{native}$. Random $128{\times}128$ patches are cropped from the
full slice.

\noindent\textbf{Data augmentation.}
Random horizontal and vertical flips (each with probability 0.5),
random intensity scaling ($\times[0.95, 1.05]$), and random intensity
shift ($[-0.04, +0.04]$).
CETA pairs share the same augmentation to maintain spatial consistency.

\subsection{Training Details}

\noindent\textbf{Optimizer.}
Both stages use AdamW ($\beta_1{=}0.9$, $\beta_2{=}0.999$) with no
weight decay. The learning rate schedule consists of a linear warmup
(5\% of total steps) followed by cosine annealing with a minimum
factor of 0.1. Gradient clipping is set to 1.0.

\subsection{Inference details}

\noindent\textbf{Sliding-window reconstruction.}
We reconstruct each 2D slice using $128\times128$ patches with 32-pixel overlap.
Overlapping predictions are fused using Gaussian window weighting to downweight patch borders and reduce stitching artifacts.

\noindent\textbf{Adaptive stepping.}
AIS determines the number of Euler integration steps from PAD:
$N = \text{clip}(\lfloor N_\text{max} \times \text{PAD} \rceil,\, N_\text{min},\, N_\text{max})$, where $N_\text{max}{=}15$ and $N_\text{min}{=}0$.
The step size is $\Delta t = 1/N$ with uniform time discretization $t_i = i/N$ for $i{=}0,\dots,N{-}1$.

%% file: supp/fig_qualitative_arbitrary_sr.tex
\begin{figure*}[t]
    \centering
    \includegraphics[width=\textwidth]{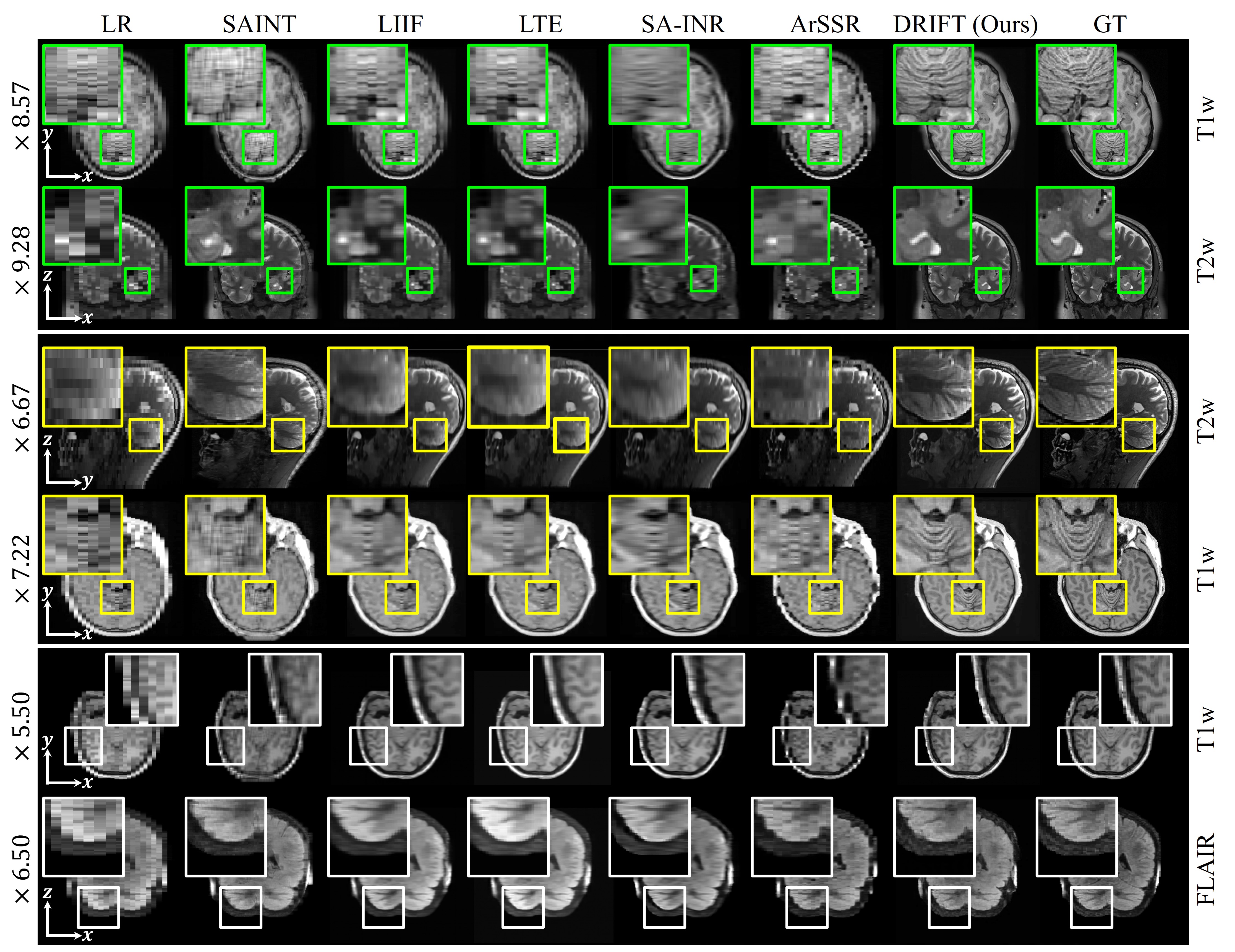}
    \caption{Qualitative comparison with arbitrary-scale SR baselines on public datasets: HCP (green), MIND (yellow), and IDEAS (white). Results are shown at representative slice-thickness settings, using HCP at $\times8.57$ (6.0\,mm) and $\times9.28$ (6.5\,mm), MIND at $\times6.67$ (6.0\,mm) and $\times7.22$ (6.5\,mm), and IDEAS at $\times5.50$ (5.5\,mm) and $\times6.50$ (6.5\,mm). Zoom-in boxes highlight regions of interest.}
    \label{fig:qual_all_arbitrary}
\end{figure*}

%% file: supp/s5_ablation_stage_quant.tex
\begin{figure}[t]
    \centering
    \includegraphics[width=\linewidth]{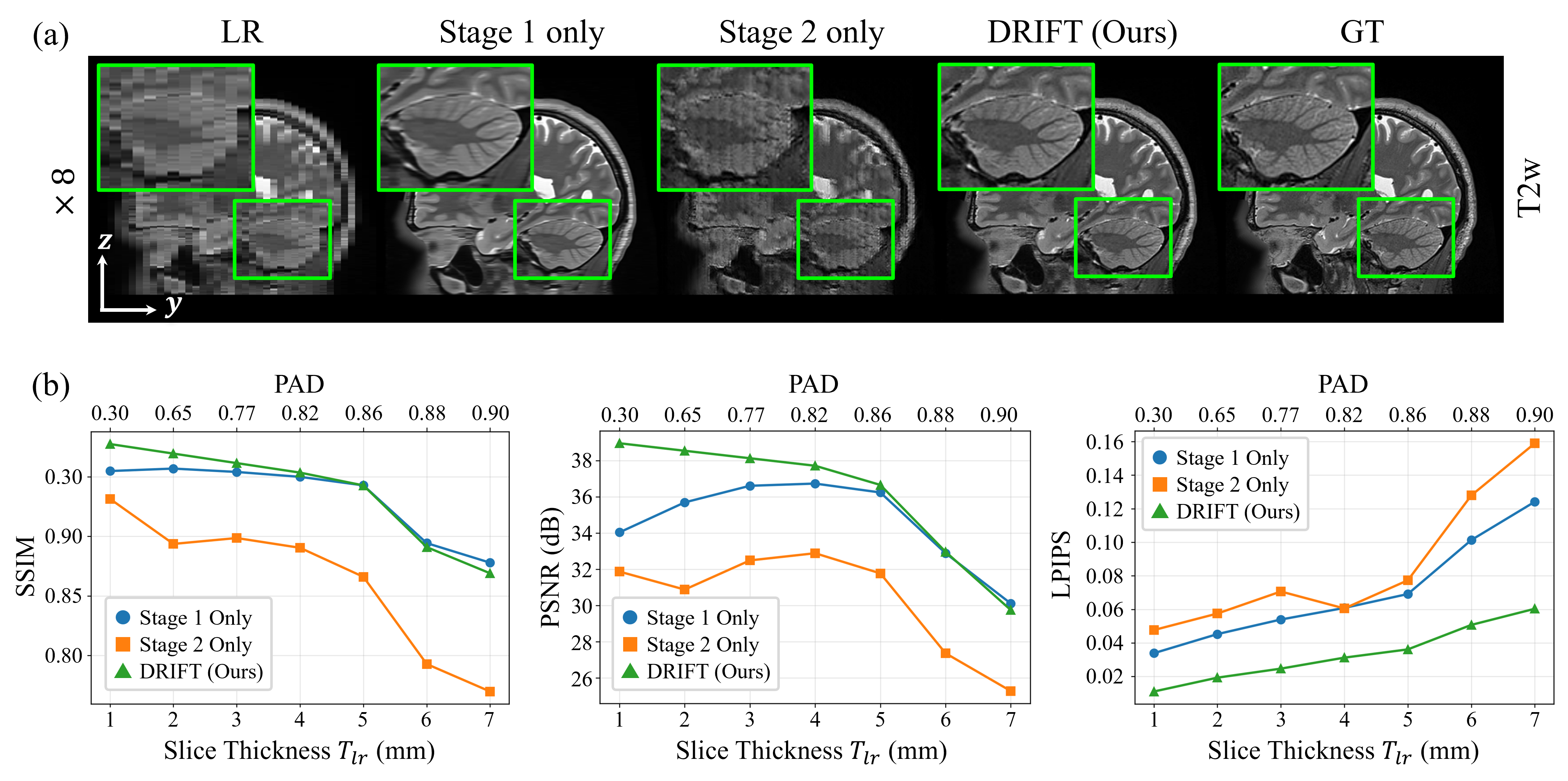}
    \vspace{-5mm}
    \caption{Stage-wise ablation of DRIFT on HCP~(T2w, $\times 8$). (a) Representative $\times 8$ qualitative comparison of LR input, Stage~1 only, Stage~2 only, full DRIFT, and GT. (b) Quantitative ablation across input slice-thicknesses. Green boxes and zoomed insets mark regions of interest. SSIM, PSNR, and LPIPS are plotted against slice-thickness $T_i$, and the top axis shows the corresponding PAD values.}
    \label{fig:supp_ablation_stage_qual_quant}
\end{figure}
\section{Stage-wise Ablation Across Slice-Thickness}
\label{sec:supp_stagewise_ablation}
Fig.~\ref{fig:supp_ablation_stage_qual_quant} provides a detailed stage-wise analysis under varying input slice thicknesses $T_i$.
We compare \emph{Stage~1 only}~(APN output), \emph{Stage~2 only}~(rectified flow without APN initialization), and full DRIFT, matching the labels in Fig.~\ref{fig:supp_ablation_stage_qual_quant}.
The qualitative example in Fig.~\ref{fig:supp_ablation_stage_qual_quant}(a) highlights distinct failure modes.
\emph{Stage~1 only} produces an anatomically coherent HR estimate, but the result is biased toward smooth structures because APN is trained as a deterministic projection from thick-slice inputs.
\emph{Stage~2 only} can sharpen local texture, but it must recover coarse anatomy and residual high-frequency detail without APN initialization, making it more sensitive to severe through-plane degradation.

The quantitative trends in Fig.~\ref{fig:supp_ablation_stage_qual_quant}(b) show consistent behavior across slice thicknesses.
At small $T_i$, \emph{Stage~2 only} partially recovers local detail, but its PSNR and SSIM decrease rapidly as $T_i$ increases, and LPIPS rises sharply at large PAD values.
\emph{Stage~1 only} varies more smoothly with thickness, reflecting the stability of deterministic anatomical projection, but it does not recover the perceptual detail of the full pipeline.
Full DRIFT first forms a stable anatomical estimate and then applies residual rectified-flow refinement, maintaining the best overall trend across PSNR, SSIM, and LPIPS as the through-plane bandwidth deficit increases.

%% file: supp/s6_ceta_weight.tex
\section{Ablation on the CETA Loss Weight}
\label{sec:supp_ablation_ceta_weights}

We ablate the weighting coefficient of the CETA term in the Stage~2 objective,
$\mathcal{L}_{\mathrm{Stage2}}=\mathcal{L}_{\mathrm{RF}}+\lambda_{\mathrm{CETA}}\mathcal{L}_{\mathrm{CETA}}$
(Sec.~3.5). Here, $\mathcal{L}_{\mathrm{RF}}$ is the primary rectified-flow supervision that matches the target velocity, whereas $\mathcal{L}_{\mathrm{CETA}}$ acts as a trajectory regularizer that encourages thickness-consistent refinement across proximal thickness pairs. We sweep $\lambda_{\mathrm{CETA}}\in\{0.1,0.3,0.5,0.7,1.0,1.5,2.0\}$ while keeping all other Stage~2 settings fixed.

\begin{figure}[t]
  \centering
  \includegraphics[width=\linewidth]{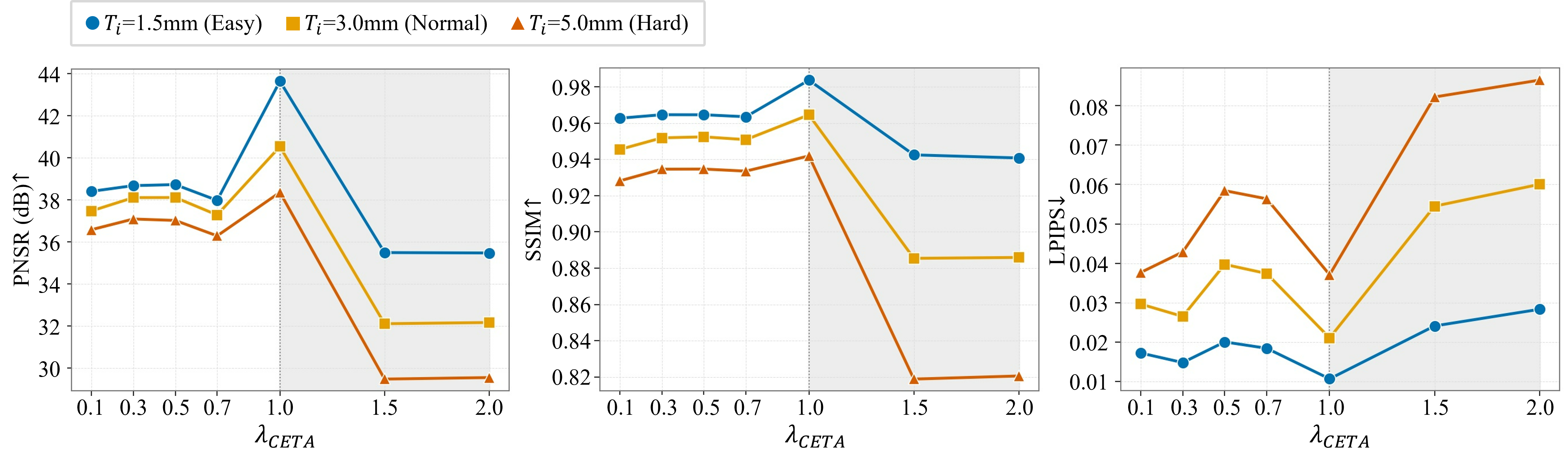}
  \caption{Ablation on the CETA weight $\lambda_{\mathrm{CETA}}$ in the Stage~2 loss $\mathcal{L}_{\mathrm{Stage2}}=\mathcal{L}_{\mathrm{RF}}+\lambda_{\mathrm{CETA}}\mathcal{L}_{\mathrm{CETA}}$. We show PSNR, SSIM, and LPIPS on HCP for $T_i\in\{1.5,3.0,5.0\}$\,mm. $T_\text{hr}$ is $0.7$\,mm.}
  \label{fig:ceta_weight}
\end{figure}

As shown in Fig.~\ref{fig:ceta_weight}, PSNR and SSIM increase as $\lambda_{\mathrm{CETA}}$ grows up to 0.5, drop slightly at 0.7, and improve again when $\lambda_{\mathrm{CETA}}$ is set to 1.0, where the best distortion performance is achieved across the evaluated thickness levels. LPIPS shows a non-monotonic but consistent improvement toward $\lambda_{\mathrm{CETA}}=1.0$: it decreases from 0.1 to 0.3, increases at 0.5, slightly decreases at 0.7, and reaches the lowest value at 1.0. Overall, setting $\lambda_{\mathrm{CETA}}=1.0$ yields the most stable and best performance across metrics, suggesting that balancing $\mathcal{L}_{\mathrm{CETA}}$ and $\mathcal{L}_{\mathrm{RF}}$ with a 1:1 ratio provides an effective trade-off between velocity matching and trajectory regularization in Stage~2.
In contrast, increasing the weight beyond 1.0 consistently lowers PSNR and SSIM and increases LPIPS, indicating that excessive CETA regularization interferes with the primary velocity-learning objective.
Accordingly, we set $\lambda_{\mathrm{CETA}}=1.0$ as the default in all experiments.

%% file: supp/s7_optimal_nfe_max.tex
\begin{figure*}[t]
    \centering
    \includegraphics[width=\textwidth]{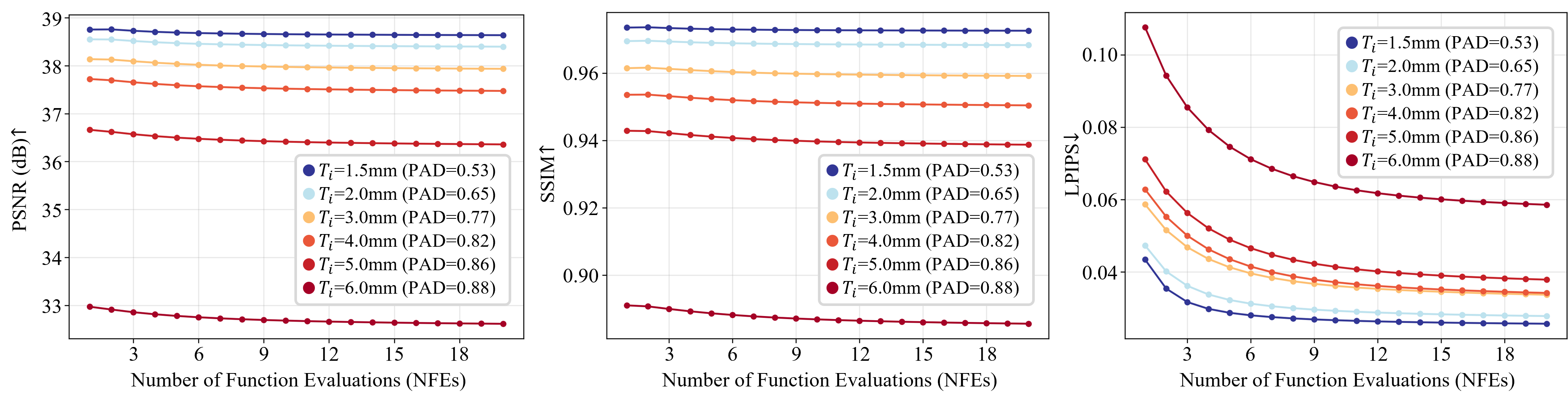}
    \caption{Effect of the NFEs in Stage~2 on HCP across slice-thickness levels. We sweep NFE from 1 to 20 and present PSNR, SSIM, and LPIPS for $T_\text{i}\in\{1.5,2.0,3.0,4.0,5.0,6.0\}$\,mm, with the corresponding PAD values.}
\label{fig:nfe_sweep}
\end{figure*}

\section{Ablation on Selecting $N_{\max}$ for Stage~2.}
\label{sec:supp_ablation_nfe_max}
We ablate NFEs in Stage~2 to select an effective $N_{\max}$ that reduces inference cost while maintaining high reconstruction quality.
Fig.~\ref{fig:nfe_sweep} shows PSNR, SSIM, and LPIPS for different NFEs under six degradation levels~(\ie, $T_i\in\{1.5,2.0,3.0,4.0,5.0,6.0\}$\,mm).
Starting from the Stage~1 initialization, increasing NFEs allows Stage~2 to progressively refine the reconstruction. As NFEs increases, PSNR and SSIM exhibit a slight decrease, which is expected from the perception--distortion trade-off in generative models~\cite{blau2018perception}. Importantly, this degradation remains minimal in our setting: even for the hardest case~($T_i=6.0$\,mm), increasing NFEs from 2 to 15 reduces PSNR by only 0.27\,dB.
In contrast, LPIPS decreases as NFEs grows, indicating that Stage~2 effectively restores fine details and improves perceptual quality beyond the coarse Stage~1 estimate.
For NFEs~$>15$, the marginal LPIPS gain per additional step becomes very small~(below $5\times10^{-4}$) across all thicknesses, suggesting saturation. We therefore set $N_{\max}=15$ as the default.

%% file: supp/s8_ablation_endpoint_bias_timestep_sampling.tex
\section{Ablation on Endpoint-Biased Timestep Sampling}
\label{sec:supp_ablation_endpoint_t}

\begin{figure}[t]
    \centering
    \includegraphics[width=0.6\textwidth]{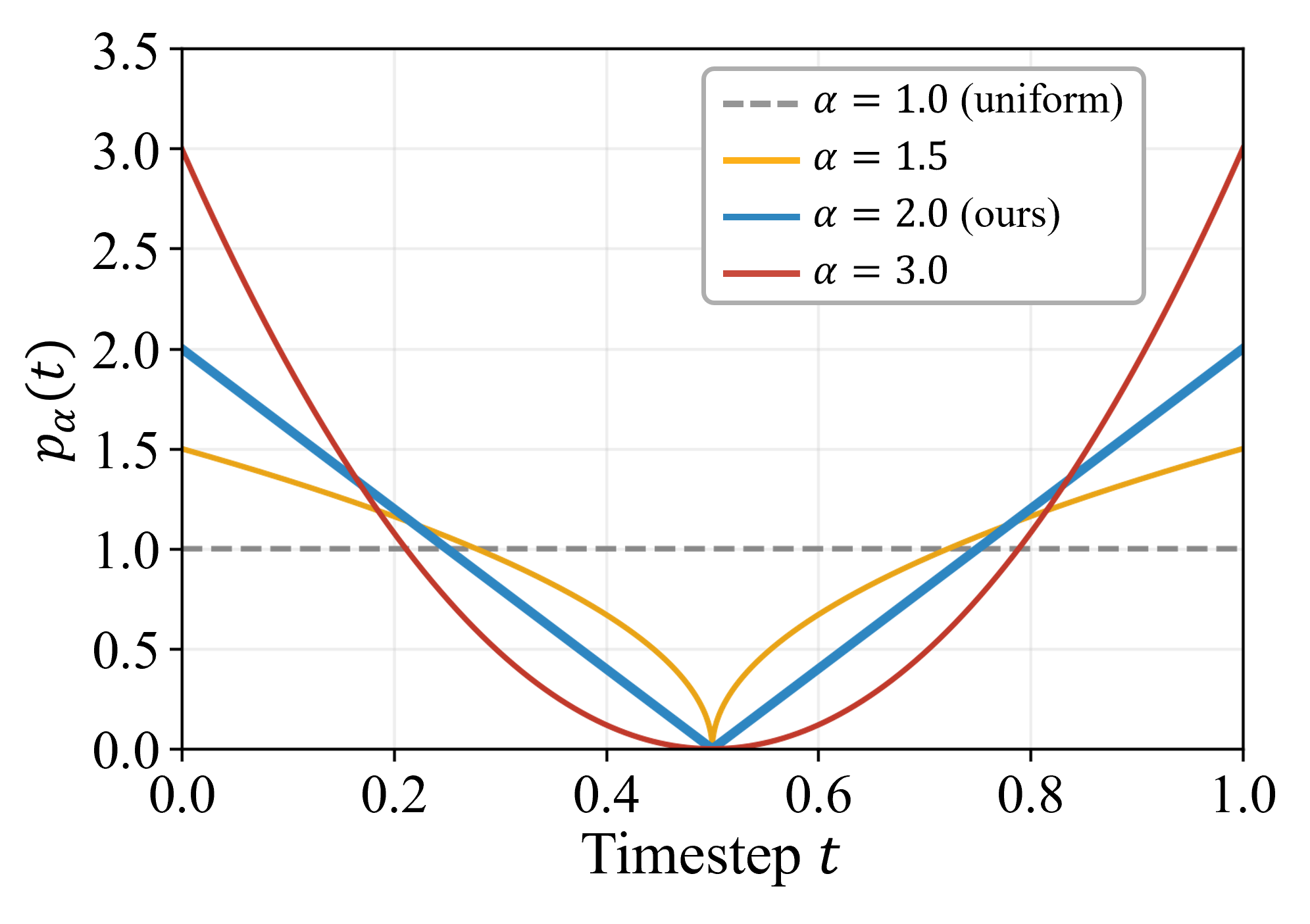}
    \caption{Endpoint-biased timestep distributions. Probability density functions $p_{\alpha}(t)$ induced by the symmetric power-law sampler~(Eq.~10) for different exponents $\alpha$. $\alpha=1$ corresponds to uniform sampling, while larger $\alpha$ places more mass near $t\approx0$ and $t\approx1$; we use $\alpha=2$ in all experiments.}
    \label{fig:ablation_endpoint_t}
\end{figure}
\begin{table}[t]
  \centering
  \caption{Sensitivity to the endpoint-biased sampling exponent $\alpha$ on HCP. $\alpha\!=\!1$ reduces to uniform sampling, while larger $\alpha$ increases probability mass near $t\!\approx\!0$ and $t\!\approx\!1$. Metrics are averaged over $T_\text{i} \in \{1.5, 3.0, 5.0\}$\,mm~($T_\text{hr}=0.7$\,mm).}
  \label{tab:ushape_sensitivity}
  \setlength{\tabcolsep}{6pt}
  \begin{tabular}{c|ccc}
    \toprule
    $\alpha$ & PSNR$\uparrow$ & SSIM$\uparrow$ & LPIPS$\downarrow$ \\
    \midrule
    1.0~(uniform) & 38.12 & 0.9458 & 0.0342 \\
    1.5            & 39.76 & 0.9568 & 0.0274 \\
    2.0     & \best{40.85} & \best{0.9634} & \best{0.0229} \\
    3.0            & 40.18 & 0.9601 & 0.0248 \\
    \bottomrule
  \end{tabular}
\end{table}

Stage~2 operates in a residual-refinement regime, where the velocity field is trained to recover high-frequency details missing from the Stage~1 estimate rather than to model the full signal.
In this setting, timestep sampling affects how training budget is distributed along the path.
Along the linear path $\mathbf{s}(t)=(1-t)\mathbf{z}+t\mathbf{y}$, mid-trajectory states are convex blends, which can attenuate sharp structures and make the high-frequency supervision less pronounced.
Conversely, the boundary regions are often more sensitive for refinement: near $t\!\approx\!0$, the model should add fine details while preserving the coarse structure of $\mathbf{z}$ and avoiding artifacts; near $t\!\approx\!1$, it should align with HR textures to mitigate over-smoothing and perceptual degradation.
This sensitivity can be more noticeable under low-budget numerical integration, where local velocity errors may accumulate along the trajectory.
Motivated by prior rectified-flow analyses suggesting that boundary timesteps can be challenging~\cite{lee2024rfpp,hu2025boundaryrfmodel}, we adopt an endpoint-biased timestep distribution $p_\alpha(t)$ induced by the symmetric power-law sampler in Eq.~10.
As illustrated in Fig.~\ref{fig:ablation_endpoint_t}, the exponent $\alpha$ provides a simple control over endpoint emphasis: $\alpha\!=\!1$ reduces to uniform sampling, while larger $\alpha$ assigns more probability mass near $t\!\approx\!0$ and $t\!\approx\!1$.
Although the corresponding density attains its minimum at $t\!=\!0.5$, this occurs at a measure-zero point, and intermediate timesteps still receive non-zero probability mass over any finite interval.
Table~\ref{tab:ushape_sensitivity} shows that moderate endpoint emphasis improves PSNR, SSIM, and LPIPS over uniform sampling on HCP, with the best performance observed at $\alpha\!=\!2.0$ in our sweep.
A larger bias~($\alpha\!=\!3.0$) yields lower scores, which suggests that overly concentrated endpoint sampling may reduce coverage of intermediate timesteps and hurt trajectory-level consistency.
Unless stated otherwise, we use $\alpha\!=\!2.0$ in all experiments.

%% file: supp/s9_ablation_ais.tex
\section{Ablation on Adaptive Step Allocation}
\label{sec:suppl_ais_ablation}

\begin{figure}[t]
    \centering
    \includegraphics[width=\linewidth]{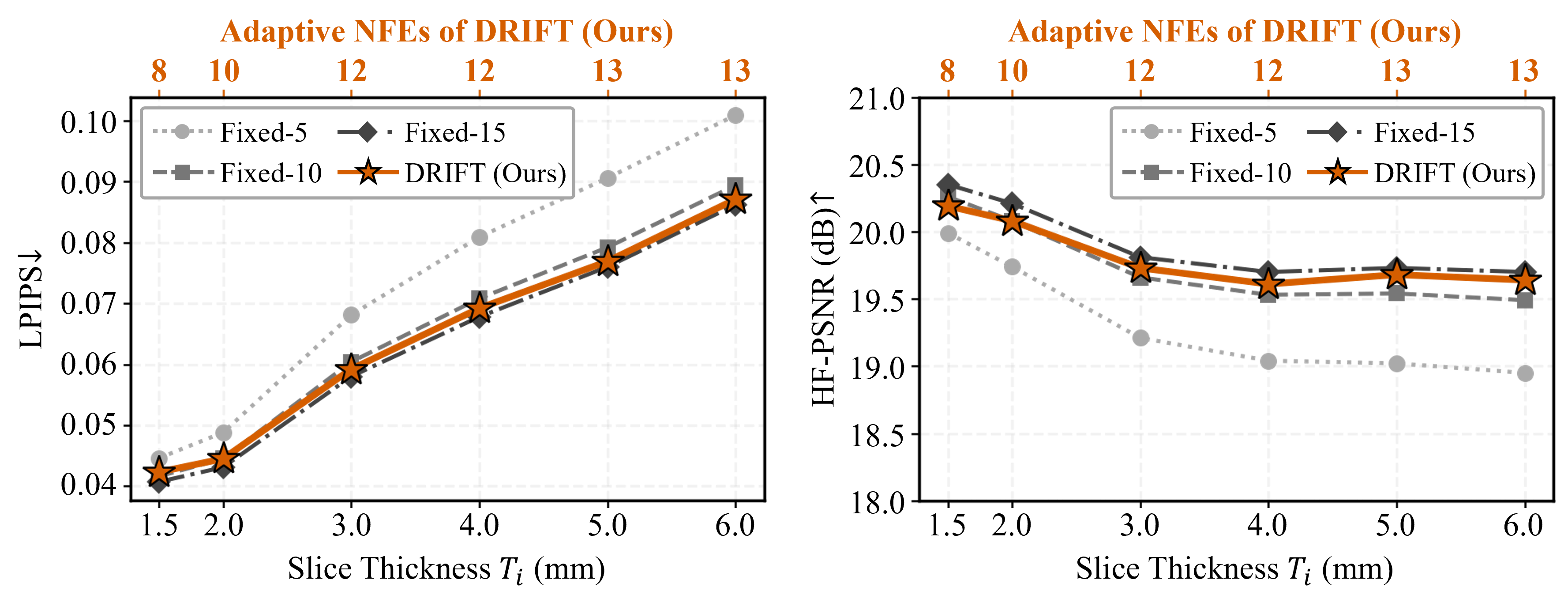}
    \caption{AIS ablation within DRIFT on HCP. LPIPS$\downarrow$~(left) and HF-PSNR$\uparrow$~(right) of the final DRIFT reconstruction across input slice-thicknesses $T_i$. Fixed-5, Fixed-10, and Fixed-15 denote DRIFT with inference steps fixed to 5, 10, and 15, respectively, while DRIFT~(Ours) uses AIS. The top axis shows the adaptive step count selected by AIS for DRIFT~(Ours) at each thickness.}
    \label{fig:ais_ablation}
\end{figure}

The AIS determines the NFEs for DRIFT from PAD, assigning fewer evaluations to thinner slices and more evaluations to thicker slices. To directly examine the standalone effect of this scheduling rule, we compare adaptive inference with fixed-step variants of the same DRIFT model. Specifically, all compared methods use the same trained Stage~1 and Stage~2 networks, and differ only in the number of inference steps: Fixed-5, Fixed-10, and Fixed-15 apply 5, 10, and 15 Euler steps uniformly across all slice-thicknesses, whereas DRIFT~(Ours) uses AIS to select the step count adaptively. We evaluate on HCP test dataset for each of six input thicknesses, $T_i \in \{1.5, 2.0, 3.0, 4.0, 5.0, 6.0\}$\,mm, with fixed target thickness $T_{\mathrm{hr}}=0.7$\,mm. Under this setting, AIS selects $\{8,10,12,12,13,13\}$ steps across the six thicknesses, corresponding to 11.3 NFEs on average. We report LPIPS and high-frequency PSNR~(HF-PSNR) to assess perceptual quality and high-frequency fidelity, respectively. As shown in Fig.~\ref{fig:ais_ablation}, adaptive DRIFT remains close to the highest fixed-step setting, Fixed-15, while using fewer evaluations on average, and consistently outperforms lower-budget fixed schedules at moderate and large thicknesses. These results support the use of AIS as an efficient inference strategy within DRIFT, allocating computation more effectively across easy and hard slice-thickness conditions.

%% file: supp/s10_ablation_pad_justification.tex
\begin{figure}[t]
    \centering
    \includegraphics[width=0.48\linewidth]{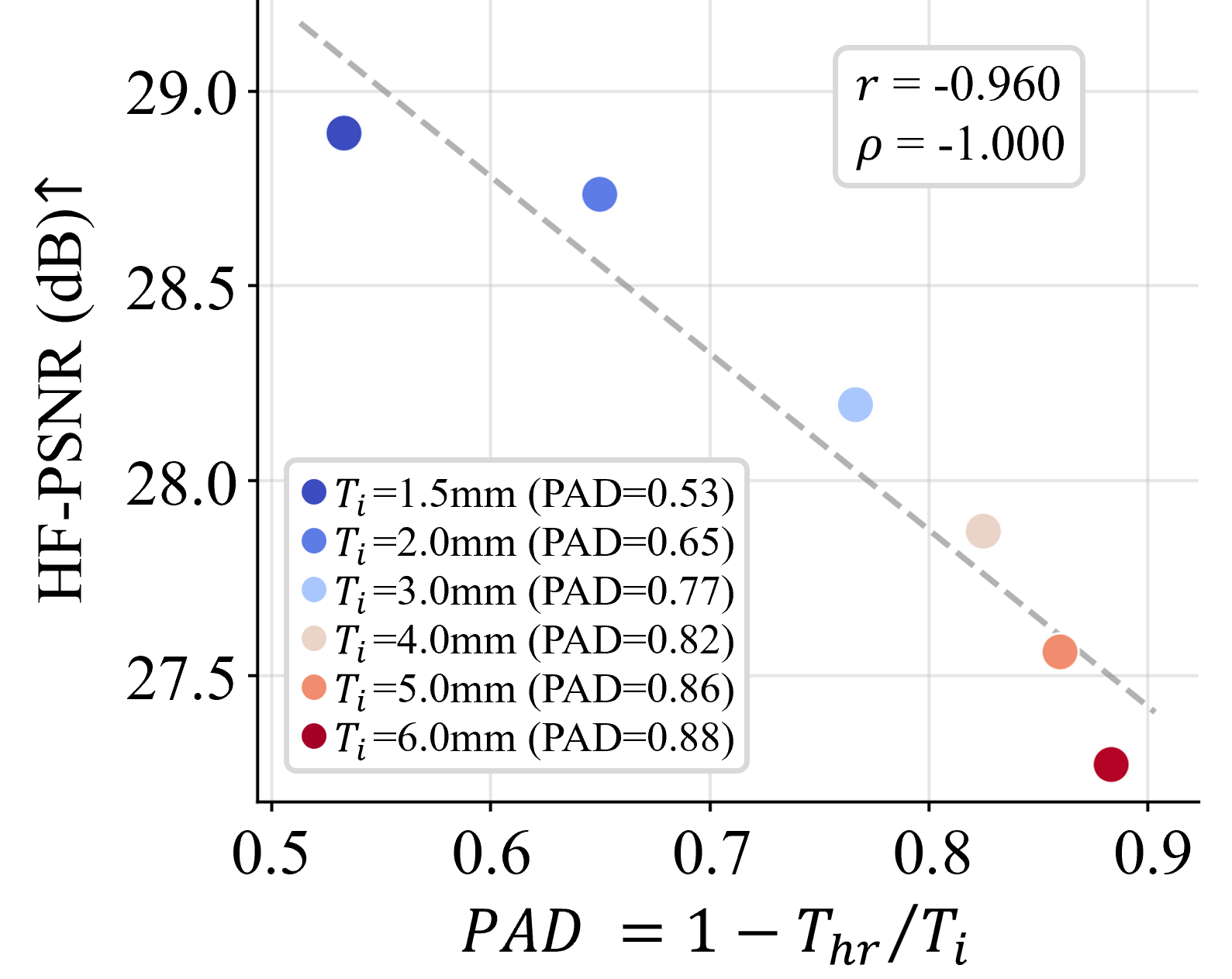}
    \caption{Correlation between PAD and reconstruction difficulty on HCP. HF-PSNR of the final DRIFT reconstruction is plotted against PAD across input slice-thicknesses $T_i$, with fixed target thickness in $T_{\mathrm{hr}}=0.7$\,mm.}
    \label{fig:pad_vs_hfpsnr}
\end{figure}

\section{Empirical Validation of PAD Against Reconstruction Difficulty}
\label{sec:suppl_pad}

The PAD is used in AIS as a metadata-driven proxy for slice-thickness difficulty. While its definition is motivated by the thickness-dependent loss of through-plane information, it remains important to examine whether PAD is consistent with empirical reconstruction difficulty across slice-thicknesses. To this end, we correlate PAD with HF-PSNR measured on the final DRIFT reconstruction. We use HF-PSNR because Stage~2 primarily refines high-frequency details that are under-recovered by Stage~1. Specifically, we apply a Laplacian high-pass filter to both the reconstruction and the HR target and compute PSNR on the filtered images. We evaluate the full DRIFT pipeline on the HCP test dataset for six input thicknesses, $T_i \in \{1.5, 2.0, 3.0, 4.0, 5.0, 6.0\}$\,mm, with fixed target thickness $T_{\mathrm{hr}}=0.7$\,mm. To isolate thickness-dependent difficulty from the effect of adaptive computation, we use the same NFEs ($N=8$) for all thicknesses. As shown in Fig.~\ref{fig:pad_vs_hfpsnr}, PAD exhibits a strong negative correlation with HF-PSNR (Pearson $r=-0.960$, Spearman $\rho=-1.000$), indicating that larger PAD values are associated with lower high-frequency reconstruction fidelity under the same inference budget. These results are consistent with the use of PAD as a simple ordering proxy for adaptive step allocation.

%% file: supp/s11_pad_vs_image_difficulty.tex
\section{PAD vs Image-Dependent Difficulty}
\label{sec:supp_pad_image_difficulty}

We evaluate whether image-dependent difficulty estimation improves adaptive step allocation beyond PAD.
Following the adaptive inference idea of AdaDiffSR~\cite{fan2024adadiffsr}, we compute an image difficulty score using an additional Stage~2 forward pass.
This experiment uses 235 T1w volumes from CC359~\cite{souza2018cc359} with multiple vendors and field strengths.
We evaluate input slice-thicknesses in $T_i \in \{1.5,2.0,3.0,4.0,5.0,6.0\}$\,mm using the IDEAS pretrained model.

Within each fixed $T_i$ group, the image difficulty score shows negligible correlation with HF-PSNR, with $|r| \leq 0.04$.
Table~\ref{tab_pad_image_difficulty} compares PAD-based scheduling with image-only and hybrid alternatives.
PAD achieves the highest aggregate PSNR without requiring an additional image difficulty forward pass.

\begin{table}[t]
\centering
\caption{Comparison of adaptive scheduling cues on CC359~\cite{souza2018cc359}.}
\label{tab_pad_image_difficulty}
\setlength{\tabcolsep}{6pt}
\begin{tabular}{lccc}
\toprule
Scheduling cue & Extra forward & Image-dependent & PSNR$\uparrow$ \\
\midrule
PAD         & \xmark & \xmark & \best{30.14} \\
Image-only  & \cmark & \cmark & 28.42 \\
Hybrid      & \cmark & \cmark & 28.38 \\
\bottomrule
\end{tabular}
\end{table}

%% file: supp/s12_ablation_ceta_timestep_sensitivity.tex
\section{Sensitivity to the CETA Proxy Timestep}
\label{sec:suppl_ceta_timestep}

We further examine the timestep used in the CETA endpoint proxy, which regularizes local cross-thickness consistency through $\tilde{\mathbf y}_{p,k}(t)=\mathbf z_{p,k}+\mathbf v_{p,k}(t)$. To assess the effect of this choice, we retrain Stage~2 while varying only the timestep used in CETA. Specifically, we compare fixed-$t$ settings with $t \in \{0.00, 0.25, 0.50, 0.75, 1.00\}$ against the current sampled-$t$ setting, which shares the same endpoint-biased timestep distribution as the rectified-flow objective. We keep the Stage~1 initialization, thickness conditioning, proximal gap $\Delta T=1.0$\,mm, $\lambda_{\mathrm{CETA}}=1.0$, and all other training settings unchanged. We evaluate on the HCP test set and average the metrics over $T_i \in \{1.5,3.0,5.0\}$\,mm with $T_{\mathrm{hr}}=0.7$\,mm.

As shown in Table~\ref{tab:ceta_t_actual}, the sampled-$t$ setting gives the highest overall performance. All fixed-$t$ variants improve over the no-CETA baseline, indicating that endpoint-consistency regularization remains beneficial even when applied at a single timestep. Among the fixed-$t$ choices, later timesteps provide stronger results than earlier timesteps, with $t=0.75$ giving the best single-$t$ performance. In contrast, $t=0.00$ gives the weakest result, suggesting that very early states provide a less informative endpoint proxy because they remain closer to the coarse Stage~1 initialization. Notably, the sampled-$t$ setting remains stronger than both exact-endpoint settings ($t=0.00$ and $t=1.00$), which suggests that CETA benefits from emphasizing boundary-adjacent regions while still maintaining coverage over a broader portion of the trajectory. Overall, these results support the current sampled-$t$ design as a simple and effective choice for CETA.

\begin{table}[t]
\centering
\caption{Sensitivity to the timestep used in the CETA endpoint proxy on the HCP test set. Stage~2 is retrained for each setting while keeping all other components fixed. Metrics are averaged over $T_i \in \{1.5,3.0,5.0\}$\,mm with $T_{\mathrm{hr}}=0.7$\,mm.}
\label{tab:ceta_t_actual}
\setlength{\tabcolsep}{6pt}
\begin{tabular}{lccc}
\toprule
CETA timestep & PSNR$\uparrow$ & SSIM$\uparrow$ & LPIPS$\downarrow$ \\
\midrule
No CETA & 36.92 & 0.9416 & 0.0308 \\
Fixed $t=0.00$ & 37.20 & 0.9440 & 0.0310 \\
Fixed $t=0.25$ & 38.00 & 0.9490 & 0.0290 \\
Fixed $t=0.50$ & 39.00 & 0.9550 & 0.0275 \\
Fixed $t=0.75$ & 39.60 & 0.9590 & 0.0255 \\
Fixed $t=1.00$ & 39.00 & 0.9560 & 0.0275 \\
Sampled $t$ (Ours) & \best{40.85} & \best{0.9634} & \best{0.0229} \\
\bottomrule
\end{tabular}
\end{table}

%% file: supp/s13_ablation_continuity.tex
\section{Through-Plane Continuity Analysis}
\label{sec:supp_continuity}

\begin{figure*}[t]
    \centering
    \includegraphics[width=\linewidth]{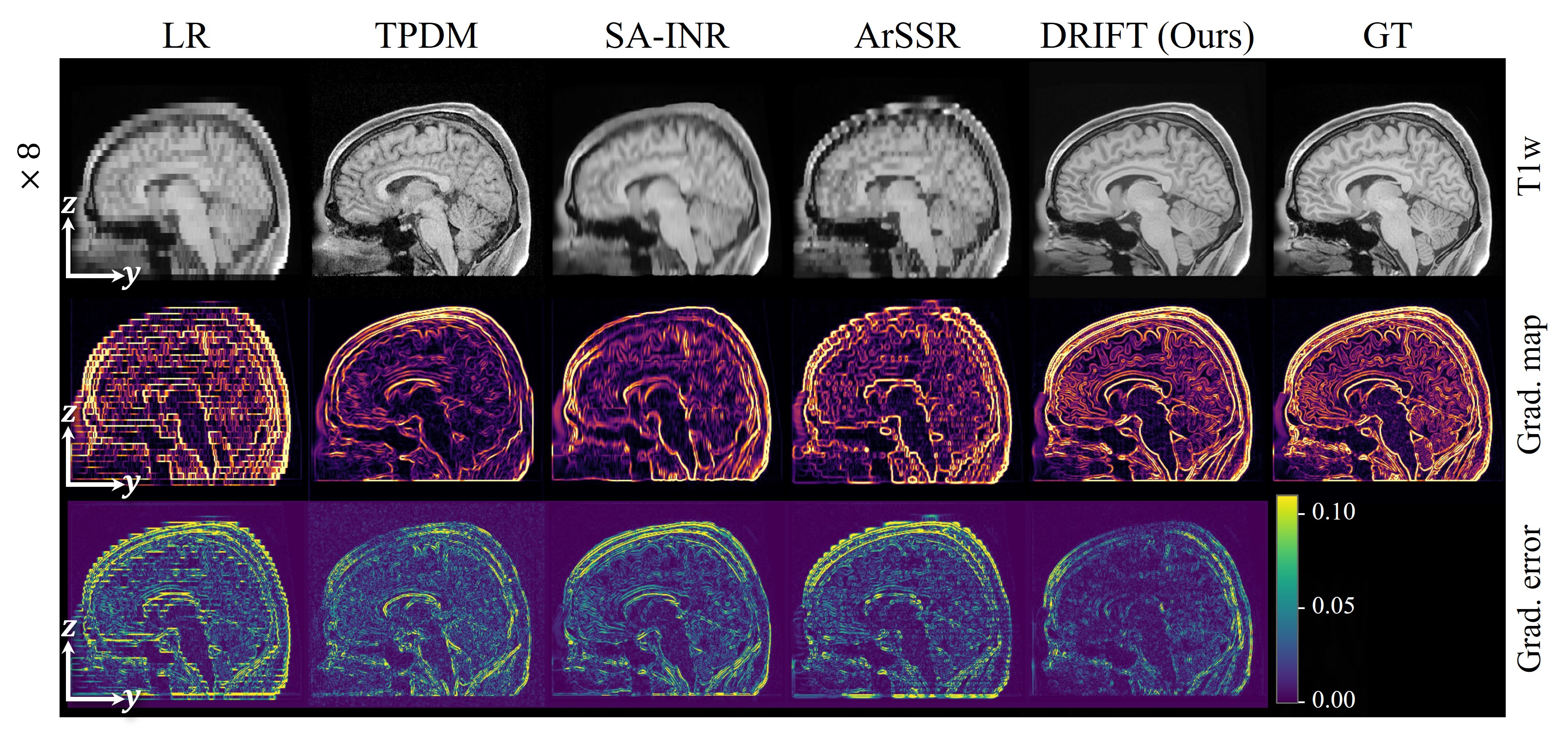}
    \caption{Through-plane continuity analysis on a representative HCP $\times 8$ case~(5.6 mm $\rightarrow$ 0.7 mm). From top to bottom: reconstructed sagittal slice, $z$-direction gradient map, and gradient-error map with respect to the ground-truth gradient map. From left to right: LR, TPDM~\cite{lee2023tpdm}, SA-INR~\cite{wang2024sainr}, ArSSR~\cite{wu2022arssr}, DRIFT, and GT.}
    \label{fig:gradmap_hcp56}
\end{figure*}

\begin{table}[t]
\centering
\caption{Through-plane continuity analysis on a representative HCP $\times 8$ case~(5.6 mm $\rightarrow$ 0.7 mm). We report Grad-RMSE between the reconstructed $z$-direction gradient map and the ground-truth gradient map. Lower is better.}
\label{tab:grad_rmse}
\setlength{\tabcolsep}{10pt}
\begin{tabular}{lc}
\toprule
Method & Grad-RMSE$\downarrow$ \\
\midrule
TPDM~\cite{lee2023tpdm} & 0.2222 \\
SA-INR~\cite{wang2024sainr} & 0.1754 \\
ArSSR~\cite{wu2022arssr} & 0.1955 \\
DRIFT~(Ours) & \best{0.1252} \\
\bottomrule
\end{tabular}
\end{table}

To complement the qualitative continuity comparisons in the main paper, we further analyze through-plane continuity on a representative HCP $\times 8$ case~(5.6 mm $\rightarrow$ 0.7 mm). We select three representative baselines from the main comparisons, TPDM~\cite{lee2023tpdm}, SA-INR~\cite{wang2024sainr}, and ArSSR~\cite{wu2022arssr}, all of which are MRI SR baselines that showed strong performance in our main experiments, and compare them with DRIFT. For each reconstructed volume $\hat{\mathbf V}$ and ground-truth volume $\mathbf V$, we compute the gradient map along the through-plane~($z$) direction and quantify the discrepancy using
\begin{equation}
\mathrm{Grad\text{-}RMSE}
=
\sqrt{\frac{1}{|\Omega|}\left\|\nabla_z \hat{\mathbf V}-\nabla_z \mathbf V\right\|_2^2},
\end{equation}
where $\Omega$ denotes the voxel grid. Lower Grad-RMSE indicates that the reconstructed slice-to-slice transition more closely follows the ground-truth anatomical variation.

As shown in Table~\ref{tab:grad_rmse}, DRIFT achieves the lowest Grad-RMSE among all compared methods. Fig.~\ref{fig:gradmap_hcp56} provides a representative visual example. Although TPDM~\cite{lee2023tpdm} reconstructs visually plausible slices, its through-plane gradient map deviates noticeably from the ground truth, and the corresponding gradient-error map contains broad residual noise across the reconstructed image. SA-INR~\cite{wang2024sainr} and ArSSR~\cite{wu2022arssr} improve over TPDM~\cite{lee2023tpdm} in Grad-RMSE, but their reconstructions remain either relatively smooth or affected by residual stair-step transitions, which appear as weakened or fragmented structures in the gradient maps. In contrast, DRIFT produces gradient patterns that more closely follow the ground truth and yields the smallest gradient-error response, particularly around fine anatomical interfaces. These observations are consistent with the qualitative findings in the main paper and support that DRIFT better preserves through-plane continuity under severe thick-slice degradation.

\begin{figure*}[t]
    \centering
    \includegraphics[width=\columnwidth]{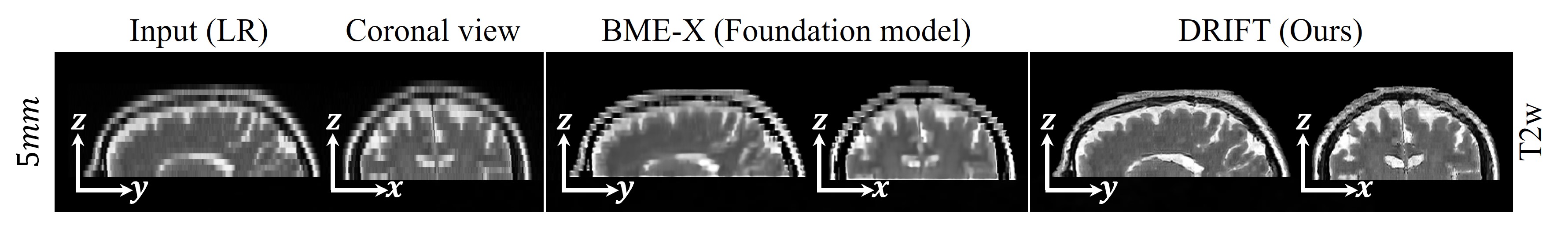}
    \caption{Zero-shot qualitative comparison results on fastMRI T2w scans.}
    \label{fig:fastmri_zeroshot_qual}
\end{figure*}

\begin{table}[t]
\centering
\caption{Zero-shot no-reference image-quality comparison on fastMRI T2w scans.}
\label{tab:fastmri_no_ref_quant}
\setlength{\tabcolsep}{10pt}
\begin{tabular}{lcc}
        \toprule
        fastMRI & \textbf{DRIFT} & BME-X~\cite{sun2025bmex} \\
        \midrule
        Sharp.$\uparrow$ & \best{0.291} & 0.238 \\
        NIQE$\downarrow$ & \best{5.30} & 7.22 \\
        BRISQ$\downarrow$ & \best{21.16} & 62.46 \\
        \bottomrule
    \end{tabular}
\end{table}

%% file: supp/s14_downstream_segmentation.tex
\section{Downstream Segmentation Evaluation}
\label{sec:supp_downstream}

We further evaluate whether through-plane SR affects downstream anatomical segmentation.
We apply SynthSeg~\cite{billot2023synthseg} to HCP T1w reconstructions under the $\times 8$ setting and compute Dice scores against segmentations obtained from the HR reference.
DRIFT achieves a Dice score of 0.910, compared with 0.816 for SA-INR, 0.801 for TPDM, and 0.772 for ArSSR.
\begin{figure*}[t]
    \centering
    \includegraphics[width=\columnwidth]{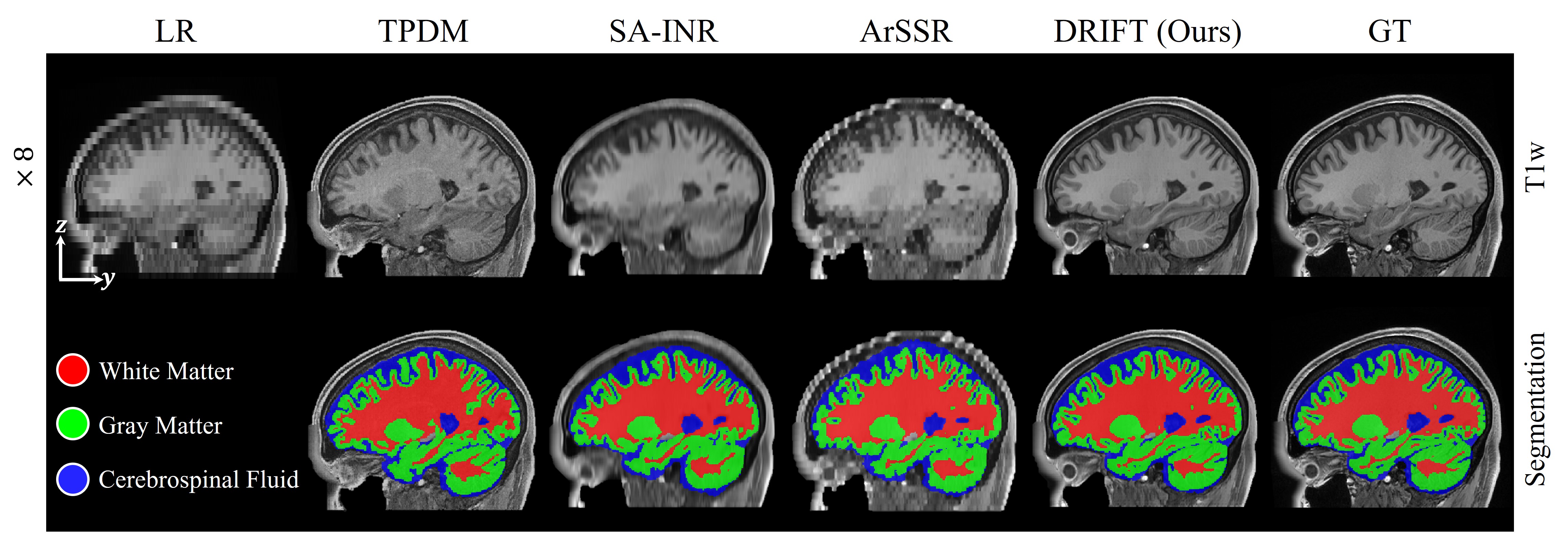}
    \caption{Downstream segmentation evaluation on HCP T1w under the $\times 8$ setting. SynthSeg is applied to each reconstructed volume, and segmentation results are compared with the HR reference.}
    \label{fig:supp_downstream_seg}
\end{figure*}
The visual comparison in Fig.~\ref{fig:supp_downstream_seg} shows the same trend, where DRIFT preserves anatomical boundaries more consistently in the reformatted view.
These results suggest that the improved through-plane reconstruction of DRIFT is also reflected in segmentation consistency, although this experiment is intended as a downstream validation rather than a task-specific optimization.

%% file: supp/s15_multiplanar_vol_visualization.tex
\section{Multi-planar Volumetric Visualization}
\label{sec:supp_multiplanar}

\begin{figure}[t]
\centering
\includegraphics[width=\linewidth]{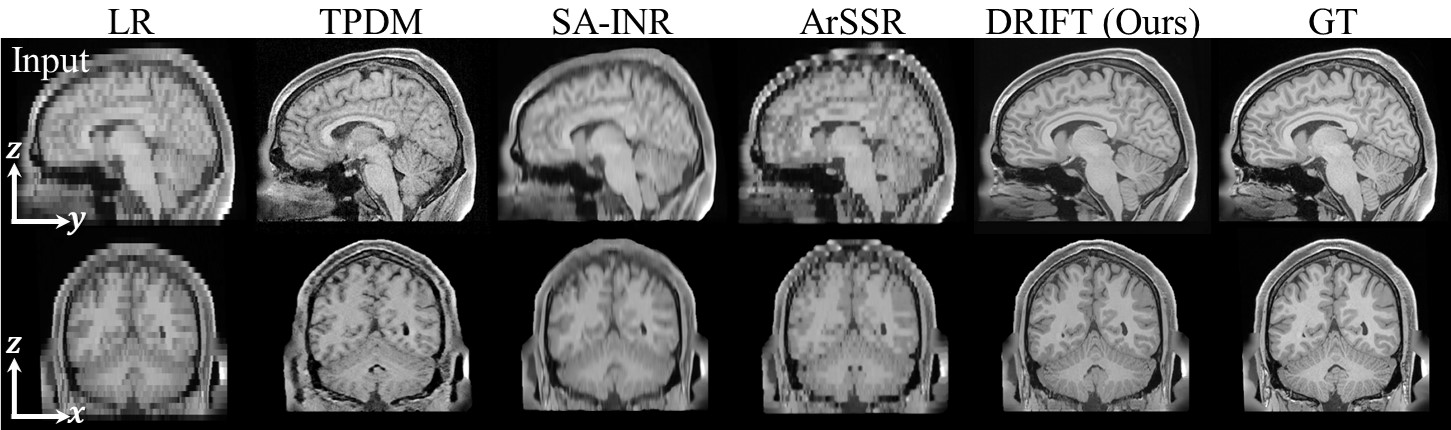}
\caption{Multi-planar volumetric visualization after through-plane SR. The upper row shows the sagittal input view, and the lower row shows the coronal reformat of the reconstructed volume.}
\label{fig:supp_multiplanar}
\end{figure}

DRIFT is implemented as a 2D slice-wise model and does not use explicit 3D convolutions or cross-slice attention.
Therefore, we do not claim a formal 3D consistency guarantee.
Instead, we examine whether independently reconstructed slices form coherent volumetric reformats in practice.
Fig.~\ref{fig:supp_multiplanar} shows a sagittal input view and a coronal reformatted view after through-plane SR.
Several baselines reduce the apparent slice spacing but still show residual stair-step artifacts, blurring, or discontinuous anatomical boundaries in the coronal reformat.
DRIFT produces a cleaner coronal reformat with more continuous cortical and ventricular structures, which is consistent with deterministic APN initialization followed by shared thickness-conditioned rectified-flow refinement across neighboring slices.
This visualization supports the practical volumetric coherence of DRIFT, while keeping the scope limited to empirical multi-planar assessment.

